\documentclass[10pt,twocolumn,letterpaper]{article}

\usepackage[pagenumbers]{iccv} 

\definecolor{iccvblue}{rgb}{0.21,0.49,0.74}
\usepackage[pagebackref,breaklinks,colorlinks,allcolors=iccvblue]{hyperref}

\def\paperID{7584} 
\def\confName{ICCV}
\def\confYear{2025}

\usepackage{bm}
\usepackage{amsmath}
\usepackage{graphicx}
\usepackage[linesnumbered, ruled, lined]{algorithm2e}
\usepackage{multirow}

\newcommand{\myPara}[1]{\vspace{3pt}\noindent\textbf{#1}}
\DeclareFixedFont{\ttb}{T1}{txtt}{bx}{n}{8} 
\DeclareFixedFont{\ttm}{T1}{txtt}{m}{n}{8}  

\usepackage{color}
\definecolor{deepblue}{rgb}{0,0,0.5}
\definecolor{deepred}{rgb}{0.6,0,0}
\definecolor{deepgreen}{rgb}{0,0.5,0}

\usepackage{listings}

\newcommand\pythonstyle{\lstset{
language=Python,
basicstyle=\ttm,
morekeywords={self},              
keywordstyle=\ttb\color{deepblue},
emph={MyClass,__init__},          
emphstyle=\ttb\color{deepred},    
stringstyle=\color{deepgreen},
frame=tb,                         
showstringspaces=false
}}

\lstnewenvironment{python}[1][]
{
\pythonstyle
\lstset{#1}
}
{}

\newcommand\pythoninline[1]{{\pythonstyle\lstinline!#1!}}

\definecolor{myRed}{rgb}{0.8,0.2,0.2}
\definecolor{myBlue}{rgb}{0.2,0.2,0.8}

\newcommand{\figref}[1]{Fig.~\ref{#1}}
\newcommand{\tabref}[1]{Tab.~\ref{#1}}
\newcommand{\secref}[1]{Sec.~\ref{#1}}

\def\sArt{state-of-the-art~}

\title{GLS: Geometry-aware 3D Language Gaussian Splatting}

\author{Jiaxiong Qiu$^{1}$
,
Liu Liu$^{1}$
,
Xinjie Wang$^{1}$
,
Tianwei Lin$^{1}$
,
Wei Sui$^{2}$
,
Zhizhong Su$^{1}$
 \\
$^{1}$Horizon Robotics, Beijing, China \quad 
$^{2}$D-Robotics, Beijing, China\\
}

\begin{document}
\maketitle
\begin{abstract}
Recently, 3D Gaussian Splatting (3DGS) has achieved impressive performance on indoor surface reconstruction and 3D open-vocabulary segmentation. This paper presents GLS, a unified framework of 3D surface reconstruction and open-vocabulary segmentation based on 3DGS. GLS extends two fields by improving their sharpness and smoothness. For indoor surface reconstruction, we introduce surface normal prior as a geometric cue to guide the rendered normal, and use the normal error to optimize the rendered depth. For 3D open-vocabulary segmentation, we employ 2D CLIP features to guide instance features and enhance the surface smoothness, then utilize DEVA masks to maintain their view consistency. Extensive experiments demonstrate the effectiveness of jointly optimizing surface reconstruction and 3D open-vocabulary segmentation, where GLS surpasses state-of-the-art approaches of each task on MuSHRoom, ScanNet++ and LERF-OVS datasets. Project webpage: \url{https://jiaxiongq.github.io/GLS_ProjectPage}. 
\end{abstract}
\section{Introduction}
\begin{figure}[t]
\includegraphics[width=\linewidth]{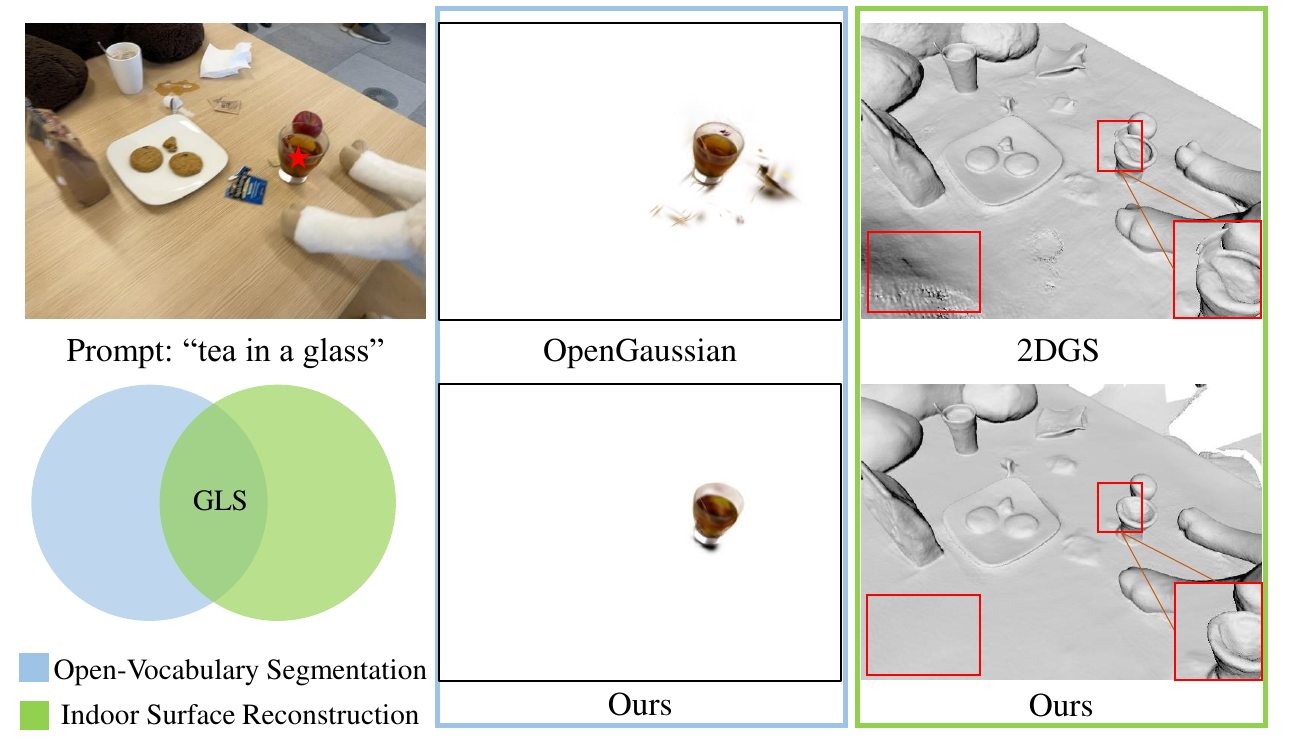} 
\vspace*{-7mm}
\caption{Indoor surface reconstruction and 3D open-vocabulary segmentation. Shadow and high-light regions make \sArt methods struggle in indoor scenes. Our proposed GLS jointly optimizes two tasks based on 3DGS and achieves much better results than OpenGaussian\cite{wu2024opengaussian} and 2DGS\cite{Huang2DGS2024}.}
\vspace*{-7mm}
\label{fig:fig1}
\end{figure}

Surface reconstruction\cite{guedon2024sugar,Huang2DGS2024,zhang2024rade,Yu2024GOF,chen2024pgsr,zhang2024gspull} and 3D open-vocabulary segmentation\cite{qin2024langsplat,shi2024language,zhou2024feature,gaussian_grouping,guo2024semantic,wu2024opengaussian}
based on 3DGS\cite{kerbl3DGaussians}, being widely applied in AR/VR\cite{jiang2024vr} and embodied intelligence\cite{botashev2024gsloc, patil2024radiance} due to its capabilities of efficient training and real-time rendering. Recently, notable works have achieved significant progress in both areas. 
For surface reconstruction, SuGaR\cite{guedon2024sugar} proposes regularization terms to align Gaussians and scene surface, then uses Possion reconstruction\cite{kazhdan2013screened} to extract mesh from Gaussians. 
For 3D open-vocabulary segmentation, LangSplat\cite{qin2024langsplat} and OpenGaussian\cite{wu2024opengaussian} successfully introduces SAM\cite{kirillov2023segany} and CLIP\cite{ilharco2021gabriel} to 3DGS. 
Gaussian grouping\cite{gaussian_grouping} utilizes a universal temporal propagation model DEVA\cite{cheng2023tracking} to obtain consistent object masks across views and propose a 3D regularization term to grouping Gaussians. 
However, these methods only focus on one task and suffer from unstable performance in complex indoor scenes as \figref{fig:fig1} shows.

In this work, we aim to achieve efficient and robust indoor surface reconstruction and 3D open-vocabulary segmentation based on 3DGS.
Our goal has two main motivations. On the one hand, the 2D open-vocabulary supervision\cite{kirillov2023segany,ilharco2021gabriel} is naturally view-inconsistent, which easily results in noise body and blur boundary of the segmented object from Gaussians. The accurate scene surface consists of sharp and smooth object surfaces, illustrating that Gaussians are mainly distributed on objects and preserve the object boundaries. This property can make the object segmentation results from Gaussians cleaner and sharper.  
On the other hand, due to the complex materials and lighting conditions of the indoor scene, 
the shadow and high-light regions on texture-less and reflective objects always result in noisy surfaces. Fortunately, accurate object masks erase the interference details on the object's surface. 
Hence, the object segmentation results can supply the smoothness prior to reducing the reconstruction noises of these objects.
In general, the optimization goals of the two tasks can be considered to be the same.

\begin{figure}[t]
\includegraphics[width=\linewidth]{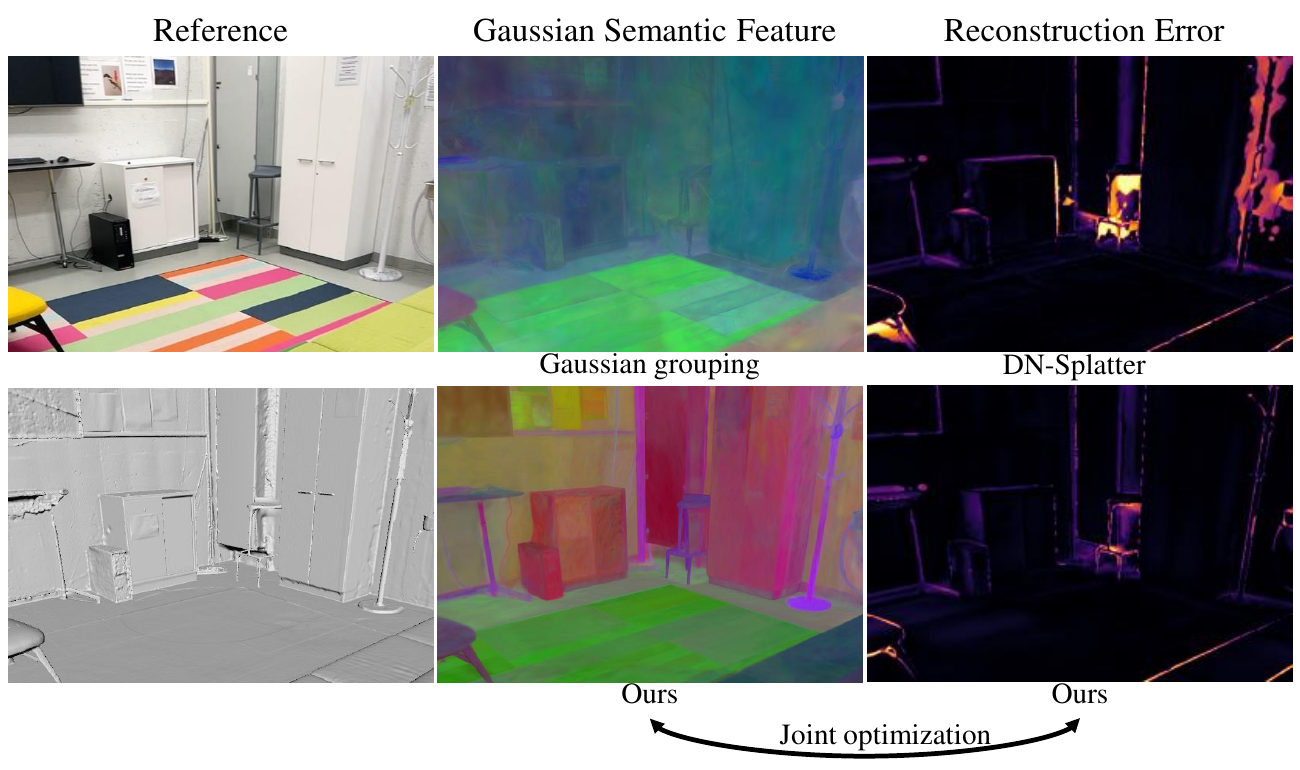}
\vspace*{-7mm}
\caption{Effect of joint optimization between indoor surface reconstruction and 3D open-vocabulary segmentation. Our method significantly improves the quality of Gaussian semantic features and surface reconstruction.}
\vspace*{-5mm}
\label{fig:fig1_2}
\end{figure}

\begin{figure*}[t]
\centering
\includegraphics[width=\textwidth]{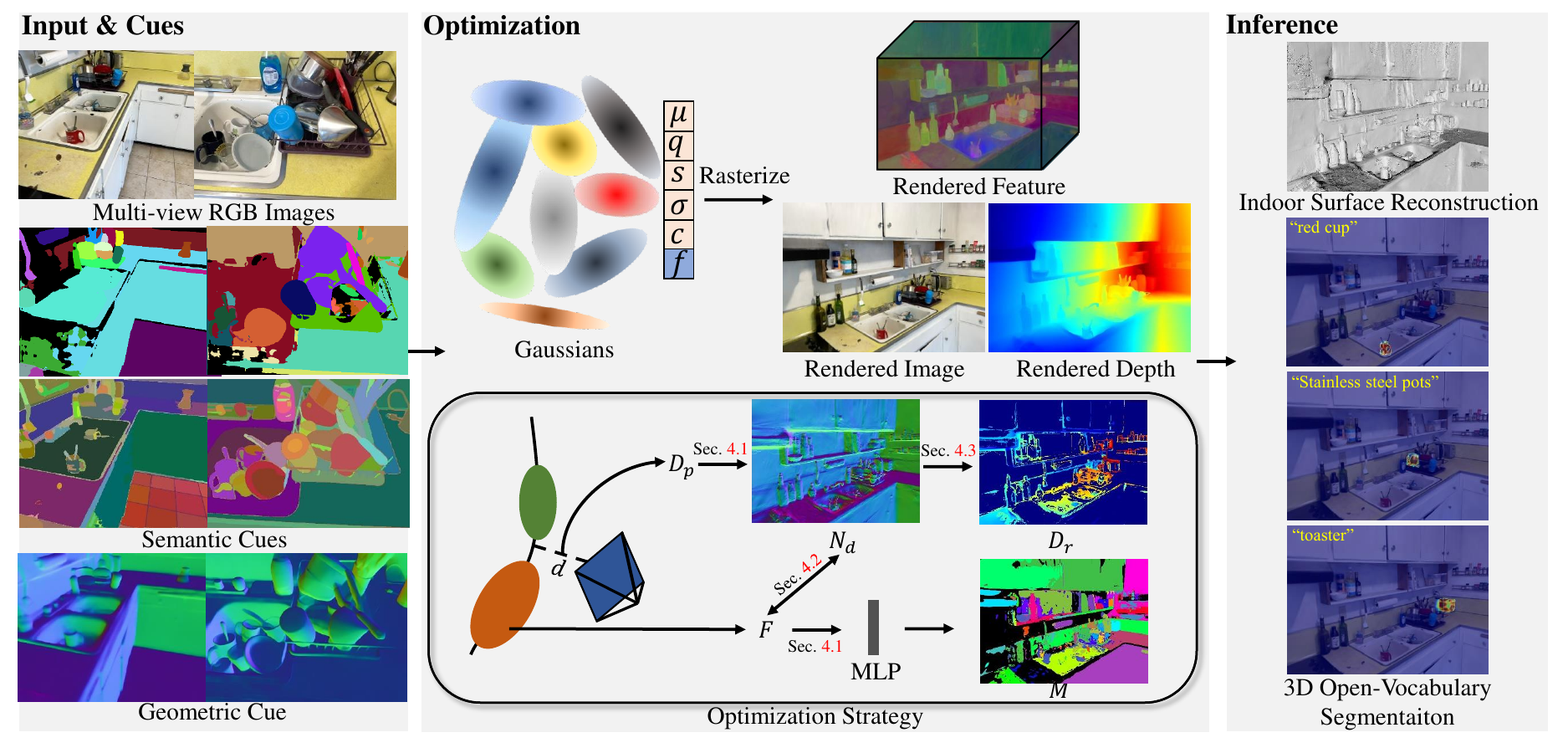}
\vspace*{-7mm}
\caption{Overview of our framework. We adopt geometric and semantic cues produced by generalizable models to jointly strengthen the reconstruction and segmentation quality of 3DGS. 
Semantic cues consist of view-consistent segmentation results of DEVA\cite{cheng2023tracking} and object-level CLIP features generated from LangSplat\cite{qin2024langsplat}. Geometric cue comes from DSINE\cite{bae2024dsine}. Our framework achieves accurate indoor surface reconstruction and 3D open-vocabulary segmentation via effective regularization terms, which faithfully enable Gaussian primitives to distribute along the object surface.}
\vspace*{-5mm}
\label{fig:pip}
\end{figure*}

Based on above motivations, we introduce GLS, which leverages complementary between surface reconstruction and 3D open-vocabulary segmentation to boost the performance
of 3DGS in two tasks. The framework is presented in \figref{fig:pip}.
Specifically, we first introduce the normal prior\cite{bae2024dsine} to regularize the surface normal estimated from rendered depth. 
Then we analyzed different situations of rendered depth under different normal errors and 
propose a regularization term to enhance the sharpness of the scene surface. 
To integrate the open-vocabulary information, we add Gaussian semantic features into vanilla 3DGS and then utilize the consistent object masks from DEVA and image features from CLIP to supervise them.
In addition, we consider the clip features as the smoothness prior to strengthen the accuracy of texture-less and reflective surfaces.
Finally, we adopt TSDF fusion\cite{newcombe2011kinectfusion} from rendered depth to extract scene mesh, then compute the similarity between input text embeddings and learned semantic embeddings to acquire object masks. 
We conduct extensive experiments on MuSHRoom\cite{ren2024mushroom}, ScanNet++\cite{yeshwanth2023scannet++} and LERF-OVS \cite{qin2024langsplat,kerr2023lerf} datasets.
As \figref{fig:fig1_2} shows, the superior performance of our model on both tasks demonstrates the effectiveness of connecting indoor surface reconstruction and 3D open-vocabulary segmentation in 3DGS. 
GLS makes the reconstructed scene surface interactive. The application demos can be visualized in the supplementary videos. In summary, our technical contributions can be listed as follows:
\begin{itemize}[leftmargin=*] 
\item[1.] We design a novel framework based on 3DGS, by jointly optimizing surface reconstruction and 3D open-vocabulary segmentation in complex indoor scenes.
\item[2.] We propose two novel regularization terms with the help of geometric and semantic cues, to facilitate the sharpness and smoothness of reconstructed scene surfaces and segmented objects. 
\item[3.] Our method inherits the training and rendering efficiency of 3DGS and achieves \sArt accuracy on surface reconstruction and 3D open-vocabulary segmentation tasks. 
\end{itemize}
 

\section{Related Work}
\label{sec:rw}
\subsection{Neural Rendering}
Neural Radiance Fields (NeRF)\cite{mildenhall2020nerf} based on implicit representations have achieved remarkable advancements in novel view synthesis. 
Some NeRF-based approaches\cite{barron2021mip,barron2022mipnerf360,barron2023zipnerf} concentrate on extending NeRF to more challenging scenes, while they are still limited by low training and rendering efficiency. Alternatively, other NeRF-based methods\cite{sun2022direct,fridovich2022plenoxels,reiser2023merf,muller2022instant,yifan2019differentiable,chen2023mobilenerf} utilize explicit representations
such as voxels, grids, point clouds and meshes. They faithfully reduce the large computational cost of neural networks. 
The recent technologies\cite{kerbl3DGaussians,Yu2023MipSplatting, liang2024analytic} of neural rendering based on alpha-blend rendering have further advanced the rendering speed and quality, by optimizing the
attributes of the explicit 3D Gaussians. 
Our framework is built upon 3DGS and focuses on optimizing it to tackle both surface reconstruction and 3D open-vocabulary segmentation tasks.

\subsection{3DGS-Based Surface Reconstruction}
Surface reconstruction from multi-view images\cite{schonberger2016pixelwise,furukawa2009accurate,godard2015multi,goel2022differentiable,ren2024agsmesh,Yu2024GOF} is a challenging task in computer vision and graphics. Recently, extracting scene surfaces from Gaussian primitives has become a popular research topic.  
SuGaR\cite{guedon2024sugar} utilizes
Poisson reconstruction\cite{kazhdan2013screened} to extract mesh from sampled Gaussians and encourages
Gaussians attach to the scene surface by several regularization terms, for enhancing reconstruction accuracy. However, due to the disorder of the Gaussians and the complexity of the scene, fitting Gaussians on geometric surfaces is challenging.
2DGS\cite{Huang2DGS2024} replaces
3D Gaussian primitives with 2D Gaussian disks for efficient training. They also introduce two regularization terms of depth distortion and normal consistency to reconstruct a smooth scene surface. 
PGSR\cite{chen2024pgsr} estimates the unbiased depth to disable the degeneration of the blending coefficient, and introduces the multi-view regularization term to achieve global-consistent reconstruction.
DN-Splatter\cite{turkulainen2024dnsplatter} and VCR-GauS\cite{chen2024vcr} leverage normal or depth priors from generalizeable models\cite{bae2024dsine,bhat2023zoedepth,fu2025geowizard}, to learn effective surfel representations.
FDS\cite{chen2024fds} introduce the optical flow into 3DGS for more accurate geometry.

Due to the lack of object attributes, these methods struggle to reconstruct sharp objects against complex lighting conditions of an indoor scene. To address these issues, we propose combining 3D open-vocabulary segmentation and surface reconstruction. In contrast with previous related works\cite{yang2023self,wu2022object,wu2023objsdfplus}, our method releases the need for ground-truth segmentation supervision and achieves highly efficient training speed.

\subsection{3DGS-Based open-vocabulary segmentation}
Recent works\cite{huang2023clip2point,xue2024ulip,zhu2023pointclip} about open-vocabulary scene understanding have significant progress by integrating 2D foundational models with 3D point cloud representations. These methods project a 3D point cloud into 2D space for zero-shot learning of aligned features. 
LERF\cite{kerr2023lerf} designs a pipeline of 3D open-vocabulary segmentation by distilling features from CLIP into NeRF.
3GS-based methods add the semantic attribute into 3D Gaussians and are supervised by 2D scene priors\cite{kirillov2023segany,ilharco2021gabriel} to overcome large computational cost of NeRF-based methods. 
LangSplat\cite{qin2024langsplat} designs a per-scene autoencoder to reduce the dimension of CLIP features on multi-level latent spaces, this scheme generates clear boundaries of rendered features.
LEGaussians\cite{shi2024language} integrate uncertainty with CLIP and DINO\cite{caron2021emerging} image features to Gaussians.  
To learn high-dimensional semantic features, Feature3DGS\cite{zhou2024feature} develops a parallel Gaussian rasterizer. 
Gaussian Grouping\cite{gaussian_grouping} introduces view-consistent masks from DEVA and proposes a 3D local consistency regularization term to strengthen the segmentation accuracy.  
OpenGaussian\cite{wu2024opengaussian} focuses on 3D point-level open-vocabulary understanding via CLIP features and proposes a two-level codebook scheme with SAM masks to refine the rendered mask.
Similar to these methods, we introduce CLIP features and SAM masks to supervise the semantic features of 3DGS.

\section{Preliminary}
\label{sec:pre}
Initialized with point clouds and colors produced by
SfM\cite{schonberger2016pixelwise}, 3DGS\cite{kerbl3DGaussians} adopt differentiable 3D Gaussian primitives $\{\bm{G}\}$ to explicitly represent a scene. Each primitive is parameterized by the Gaussian function:
\begin{equation}
\setlength{\abovedisplayskip}{3pt}
\setlength{\belowdisplayskip}{3pt}
\bm{G}(\bm{x}|\bm{\mu}, {\textstyle \tiny \bm{\sum}}) = e^{-\frac{1}{2}(\bm{x}-\bm{\mu})^T\bm{\sum}^{-1}(\bm{x}-\bm{\mu})},
\label{eq:Gaussian}
\end{equation}
where $\bm{\mu} \in \mathbb{R}^3$ and ${\textstyle \tiny \bm{\sum}} \in \mathbb{R}^{3 \times 3}$ are the center and the covariance matrix of spatial points $\bm{x}$ respectively. ${\textstyle \tiny \bm{\sum}}$ consists of a scaling vector $\bm{s} \in \mathbb{R}^3$ and a quaternion vector $\bm{q} \in \mathbb{R}^4$.

3DGS enables an efficient alpha-blending procedure for real-time rendering. Given a camera view, 3D Gaussian primitives are projected into viewing space to be 2D Gaussians, which are sorted by the z-buffer strategy and alpha-composited by the volume rendering equation\cite{mildenhall2020nerf} to generate pixel colors $\bm{C}$: 
\begin{equation}
\setlength{\abovedisplayskip}{3pt}
\setlength{\belowdisplayskip}{3pt}
\bm{C} = \sum_{i \in N}\bm{c}_i\alpha_i\prod_{j=1}^{i-1}(1-\alpha_j),
\label{eq:volumn_rendering}
\end{equation}
where $N$ is the number of 3D Gaussian primitives, $\bm{c} \in \mathbb{R}^3$ is the color of a Gaussian primitive estimated from
spherical harmonics and the viewing direction. $\alpha$ 
is the blending coefficient determined by the
opacity $\sigma$. Similar to rendered color $\bm{C}$, the rendered alpha $A$, depth $D$ and rendered semantic features $\bm{F}$ can be denoted by:
\begin{equation}
\setlength{\abovedisplayskip}{3pt}
\setlength{\belowdisplayskip}{3pt}
\begin{aligned}
A &= \sum_{i \in N}\alpha_i\prod_{j=1}^{i-1}(1-\alpha_j),\\
D &= \sum_{i \in N}d_i\alpha_i\prod_{j=1}^{i-1}(1-\alpha_j),\\
\bm{F} &= \sum_{i \in N}\bm{f}_i\alpha_i\prod_{j=1}^{i-1}(1-\alpha_j).
\end{aligned}
\label{eq:volumn_rendering2}
\end{equation}
where $d$ is the distance between the 2D Gaussian point and the camera center, $\bm{f}$ is the semantic feature of each Gaussian.

\section{Method}
\label{sec:method}
Given multi-view RGB images captured by a camera in an indoor scene, our goal is to jointly reconstruct the scene and open-vocabulary objects. To achieve this goal, we introduce GLS, a novel framework based on 3DGS. As shown in\figref{fig:pip}, our framework consists of three procedures. 
In the input procedure, we use the generalizable model SAM\cite{kirillov2023segany}, DEVA\cite{cheng2023tracking} and CLIP\cite{ilharco2021gabriel} to produce 2D consistent semantic masks $\hat{M}$ and object-level features $\hat{F}$. 
Then we adopt the generalizable model\cite{bae2024dsine} of surface normal estimation to acquire the geometric cue $\hat{N}$.
In the optimization procedure, we utilize the semantic and normal priors for regularization. We first follow previous approaches to regularize the rendered color, depth and semantic feature (\secref{sec:sgc0}). Then we propose a novel smoothness term (\secref{sec:sgc2}) to tackle texture-less regions and a novel constraint by analyzing the normal error of Gaussians (\secref{sec:sgc1}) to refine object structures.
In the inference procedure, our model reconstructs the indoor surface and selects the target object by the open-vocabulary text simultaneously.

\subsection{Leveraging 2D Semantic and Geometric Cues}
\label{sec:sgc0}
Previous works\cite{turkulainen2024dnsplatter,wu2024opengaussian} either only consider surface reconstruction, or only perform the 2D open-vocabulary segmentation. We propose combining two tasks to jointly reconstruct the scene surface and segment objects of the scene. 

For indoor surface reconstruction, in contrast to DN-splatter\cite{turkulainen2024dnsplatter}, we utilize the TSDF Fusion\cite{lorensen1998marching,newcombe2011kinectfusion} to extract the mesh. Hence, we focus on optimizing the rendered depth $D$. 
As demonstrated in 2DGS\cite{Huang2DGS2024}, the local smoothness of rendered depth is challenging and important for the surface reconstruction of 3DGS. To smooth rendered depth, we introduce the normal prior $\hat{N}$. We follow the manner of PGSR\cite{chen2024pgsr} to estimate the gradients of 3D points projected from rendered depth $D$, and take them as the local surface normal $N_d$. Then we leverage $\hat{N}$ to regularize it by:
\begin{equation}
\setlength{\abovedisplayskip}{3pt}
\setlength{\belowdisplayskip}{3pt}
\mathcal{L}_n = \sum_{i,j}A(1-N_d^T\hat{N}),
\label{eq:depth2normal_loss}
\end{equation}

For 3D open-vocabulary segmentation, we directly use the view-consistent segmentation results of DEVA\cite{cheng2023tracking} as the supervision of the rendered mask $M$. Specifically, given the rendered features $F$, we use an MLP layer to increase its feature dimension to the number of total categories first. The softmax function and standard cross-entropy loss $\mathcal{L}_m$ are adopted for final classification, which is defined as: $\mathcal{L}_m = - \sum_{i=1}^{S} y_i \log(\hat{y}_i)$, where $S$ is the number of classes, $y_i$ is the true label (0 or 1) for the $i$-th class, and $\hat{y}^i$ is the predicted probability for the $i$-th class. To enhance the open-vocabulary capabilities of our model, we employ the CLIP features $\hat{F}$ to supervise $F$ by:
\begin{equation}
\setlength{\abovedisplayskip}{3pt}
\setlength{\belowdisplayskip}{3pt}
\mathcal{L}_{clip} = || F - \hat{F} ||^1,
\label{eq:clip_loss}
\end{equation}

\subsection{Semantic-feature Guided Normal Smoothing}
\label{sec:sgc2}
Big surfaces like floors and desktops are generally over-smoothing in an indoor scene. Although $\mathcal{L}_n$ supply local smoothness for reconstruction, it still struggles in shadow and high-light regions of these surfaces. High-weight $\mathcal{L}_n$ causes an over-smoothing effect and then makes some small objects disappear, while low-weight $\mathcal{L}_n$ disables the local smoothness of rendered depth. 

To resolve this issue, we propose introducing clip features to help smooth big surfaces, then we can employ $\mathcal{L}_n$ with low weight for protecting small object boundaries. 
Essentially, clip features supply high-dimension object-level smoothness which can implicitly reduce the inference of big surfaces. 
Concretely, we first introduce SAM to select objects that occupy the top-$k$ area by the mask $M_o$. and design a novel regularization term $\mathcal{L}_s$ for smoothing them as:
\begin{equation}
\setlength{\abovedisplayskip}{3pt}
\setlength{\belowdisplayskip}{3pt}
\mathcal{L}_s = M_o(|\partial_x N_d|e^{-||\partial_x \bm{\hat{F}}||^1} + |\partial_y N_d|e^{-||\partial_y \bm{\hat{F}}||^1}),
\label{eq:smooth_loss}
\end{equation}

\subsection{Normal-error Guided Depth Refinement}
\label{sec:sgc1}
\begin{figure}[t]
\centering
\includegraphics[width=\linewidth]{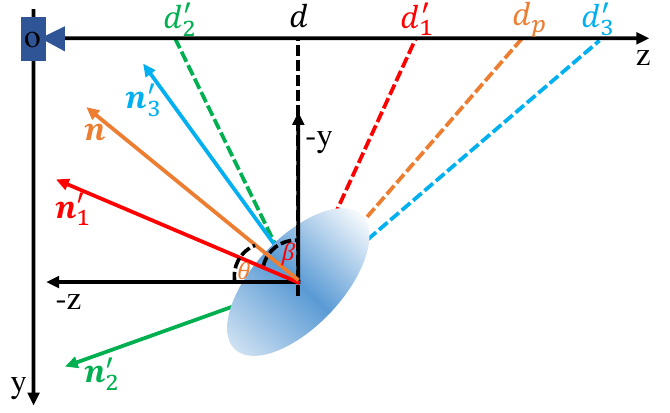}
\vspace*{-7mm}
\caption{Depth refinement guided by the normal error between the rendered normal $\bm{n}$ and the ideal normal $\bm{n}'$. We analyze three conditions of $\bm{n}'$. The ideal depth $d'$ of each condition is the unbiased depth proposed by PGSR\cite{chen2024pgsr}.}
\vspace*{-5mm}
\label{fig:a1}
\end{figure}

We take the unbiased depth $D_p$ from PGSR\cite{chen2024pgsr} to replace the rendered depth and as the input of TSDF fusion procedure, which denoted by: $D_p = \frac{D}{cos(\theta)}$, where $\theta$ is the angle between the direction of the intersecting ray and the rendered normal $\bm{n}$\cite{cheng2024Gaussianpro}. However, due to the ambiguity of the shortest Gaussian axis, $\bm{n}$ always occurs large error when compared to the ideal normal $\bm{n}'$. As \figref{fig:a1} shows, given $\bm{n}$ and the corresponding unbiased depth $d_p$, there are three main conditions of the ideal normal $\bm{n}'$ as follows: 
\begin{itemize}[leftmargin=*] 
\item[1.] $\bm{n}'_{1}$ exists between $\bm{n}$ and $\bm{-z}$. The corresponding ideal depth $d'_{1} \in [d, d_p]$.
\item[2.] $\bm{n}'_{2}$ exists between $\bm{-z}$ and $\bm{y}$. The corresponding ideal depth $d'_{2} \in [0, d]$.
\item[3.] $\bm{n}'_{3}$ exists between $\bm{n}$ and $\bm{-y}$. The corresponding ideal depth $d'_{3} \in [d_p, d]$. 
\end{itemize}

To determine the $i$-th $\bm{n}'$, we propose using the angle $\alpha$ between $\bm{n}'_i$ and $\bm{-y}$ along with $\theta' = 90^\circ - \theta$. Then we have $cos(\alpha) = \bm{n}'\cdot\bm{-y}$ and $cos(\theta') = \bm{n}\cdot\bm{-y}$. Consequently, we can get $\bm{n}'_{1} \in \{\bm{n}'|cos(\alpha) \textgreater cos(\theta') \textgreater 0\}$, $\bm{n}'_{2} \in \{\bm{n}'| cos(\alpha) \textless 0\}$ and $\bm{n}'_{3} \in \{\bm{n}'|0 \textless cos(\alpha) \leq cos(\theta')\}$. 

\begin{figure}[t]
\includegraphics[width=\linewidth]{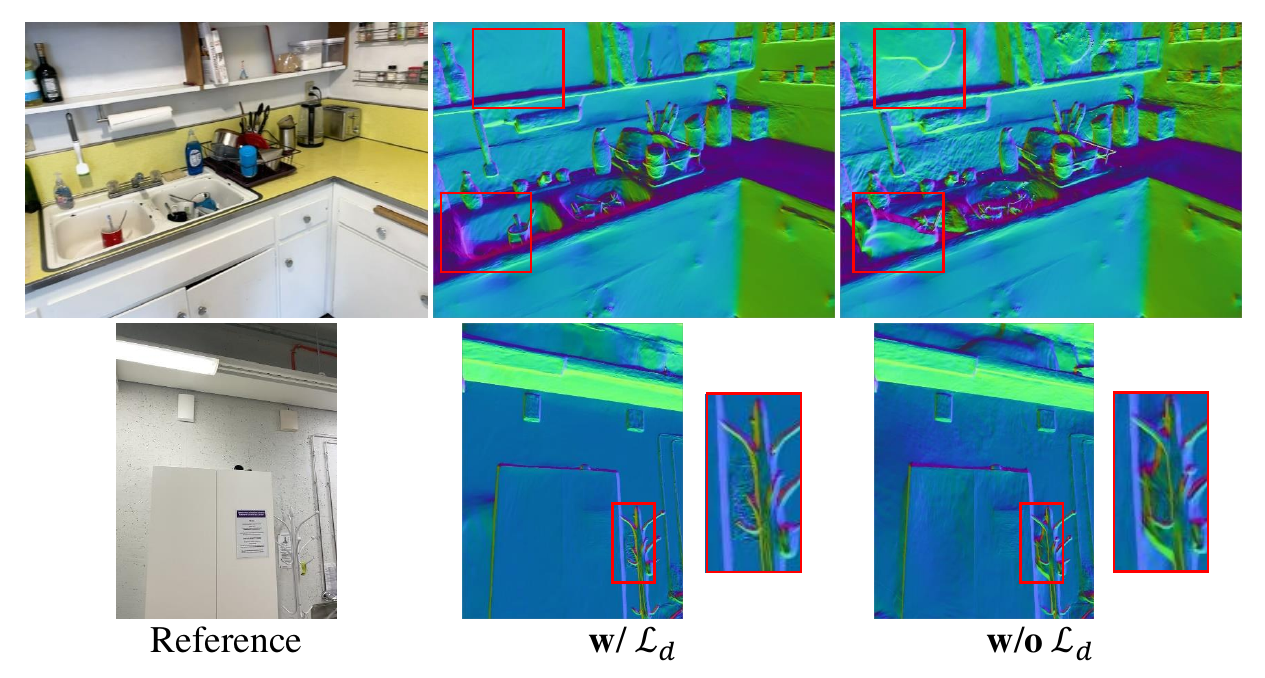} 
\vspace*{-7mm}
\caption{Effect of $\mathcal{L}_d$. The surface normal is estimated from the unbiased depth. $\mathcal{L}_d$ faithfully enhances the sharpness and smoothness of the reconstructed surface.}
\vspace*{-7mm}
\label{fig:a2}
\end{figure}

In practice, we take $\hat{N}$ to approximate $\bm{n}'$ in 2D space. Then we can acquire three masks ($M_1$,$M_2$ and $M_3$) through the above three conditions. Ulteriorly, to make $d_p$ close to $d'$, we refine $D_p$ by reconstituting depth in each mask. Specifically, for the first condition, we introduce the rendered alpha $A$ to integrate $D$ and $D_p$ in $M_1$ by: $M_1(AD_p + D - AD)$. For the second condition, we choose $M_2D$ as the target depth. For the last condition,  we choose $M_3D_p$ as the target depth. Finally, the whole target depth $D_r$ can be denoted by:
\begin{equation}
\setlength{\abovedisplayskip}{3pt}
\setlength{\belowdisplayskip}{3pt}
D_r = M_1(AD_p + D - AD) + M_2D + M_3D_p,
\label{eq:target_depth}
\end{equation}
Then we conduct a novel regularization term $\mathcal{L}_d$ between $D_r$ and $D_p$ by:
\begin{equation}
\setlength{\abovedisplayskip}{3pt}
\setlength{\belowdisplayskip}{3pt}
\mathcal{L}_d = 1 - e^{-||D_p-D_r||^1},
\label{eq:opt_d_loss}
\end{equation}
Furthermore, we choose the pixels under $\{N_d^T\hat{N} \textless 0.9\}$ to compute the value of $\mathcal{L}_d$. As \figref{fig:a2} demonstrates, $\mathcal{L}_d$ encourages the Gaussians closely attach to the object surface and then improves the sharpness and smoothness of the reconstructed surface. 

\subsection{Optimization}
\label{sec:opt}
We adopt the photometric loss $L_c$ from vanilla 3DGS\cite{kerbl3DGaussians}. 
All loss functions are simultaneously optimized by training from scratch. The total loss function $\mathcal{L}$ can be defined as:
\begin{equation}
\label{eq:total_loss}
\setlength{\abovedisplayskip}{3pt}
\setlength{\belowdisplayskip}{3pt}
\mathcal{L} = \mathcal{L}_c + \alpha_n \mathcal{L}_n + \alpha_m \mathcal{L}_m + \alpha_{clip} \mathcal{L}_{clip} + \alpha_d \mathcal{L}_d + \alpha_s \mathcal{L}_s.
\end{equation}
We set $\alpha_n = 0.07$, $\alpha_m = 0.3$, $\alpha_{clip} = 1.0$, $\alpha_d = 0.01$ and $\alpha_s = 0.5$ in our experiments.

\begin{table*}[t]
\footnotesize
\centering
\caption{Quantitative results of indoor surface reconstruction on MuSHRoom dataset (only contains: `coffee\_room', `computer', `honka', `kokko' and `vr\_room'). Our method outperforms other methods on most metrics. The best metrics are \textbf{highlighted}.}
\vspace*{-3mm}
\resizebox{\linewidth}{!}{\begin{tabular}{c|c|cccccc}
\cline{1-8}
 Methods & Sensor Depth  & Accuracy$\downarrow$ &    Completion$\downarrow$ & Chamfer$-L_1$$\downarrow$ & Normal Consistency$\uparrow$ & F-score$\uparrow$ & Time \\
\cline{1-8}
DN-Splatter\cite{turkulainen2024dnsplatter} & \checkmark  & 0.0402 & \textbf{0.0211} & 0.0306 & 0.8449 & 0.8590 & 1.0h\\
Ours & \checkmark  & \textbf{0.0288} & 0.0254 & \textbf{0.0271} & \textbf{0.8640} & \textbf{0.8924} & 0.7h \\
\cline{1-8}
2DGS\cite{Huang2DGS2024} & $\times$ & 0.1728 & 0.1619 & 
0.1674 & 0.7113 & 0.3433 & 0.3h \\
PGSR\cite{chen2024pgsr} & $\times$  & 0.4957 & 0.7672 & 0.6315 & 0.5566 & 0.1178 & 0.7h\\
FDS\cite{chen2024fds} & $\times$  & 0.0639 & \textbf{0.0528} & 0.0584 & 0.8169 & \textbf{0.6998} & 1.3h\\
Ours & $\times$  & \textbf{0.0538} & 0.0582 & \textbf{0.0560} & \textbf{0.8357} & 0.6922 & 0.7h\\
\cline{1-8}
\end{tabular}}
\label{tab:mush}
\end{table*}

\begin{table*}[t]
\footnotesize
\centering
\caption{Quantitative results of indoor surface reconstruction on ScanNet++ dataset (only contains: `8b5caf3398' and `b20a261fdf'). The best metrics are \textbf{highlighted}.}
\vspace*{-3mm}
\resizebox{\linewidth}{!}{\begin{tabular}{c|c|cccccc}
\cline{1-8}
 Methods & Sensor Depth & Accuracy$\downarrow$ &    Completion$\downarrow$ & Chamfer$-L_1$$\downarrow$ & Normal Consistency$\uparrow$ & F-score$\uparrow$ & Time\\
\cline{1-8}
DN-Splatter\cite{turkulainen2024dnsplatter} & \checkmark & 0.0977 & 0.0431 & 0.0704 & 0.8272 & 0.7094 & 1.0h\\
Ours & \checkmark & 0.0640  & \textbf{0.0272} & \textbf{0.0444} & 0.9064 & \textbf{0.8623} & 0.6h \\
\cline{1-8}
2DGS\cite{Huang2DGS2024} & $\times$ & 0.2440 & 0.4362 & 0.3401 & 0.6343 & 0.1838 & 0.2h \\
PGSR\cite{chen2024pgsr} & $\times$ & 0.1670 & 0.2188 & 0.1929 & 0.7622 & 0.2227 & 0.6h \\
Ours & $\times$ & \textbf{0.0861} & \textbf{0.1009} & \textbf{0.0935} & \textbf{0.8578} & \textbf{0.4799} & 0.6h\\
\cline{1-8}
\end{tabular}}
\vspace*{-3mm}
\label{tab:scanpp}
\end{table*}

\section{Experiments}
\label{sec:exp}
\subsection{Settings}
\paragraph{Datasets \& Metrics.}
For 3D open-vocabulary segmentation, we follow LangSplat\cite{qin2024langsplat} to conduct experiments on the LERF-OVS dataset\cite{kerr2023lerf}. We adopt the evaluation metrics from Gaussian Grouping. Specifically, we use the text query to select 3D Gaussians, then calculate the average IoU (mIoU) and boundary IoU (mBIoU) accuracy between rendered masks and annotated object masks. For indoor surface reconstruction, we conduct experiments on two RGBD datasets: MuSHRoom\cite{ren2024mushroom} (only contains: `coffee\_room', `computer', `honka', `kokko' and `vr\_room') and ScanNet++\cite{yeshwanth2023scannet++} (only contains: `8b5caf3398' and `b20a261fdf'). We and the same tool of DN-Splatter\cite{turkulainen2024dnsplatter} to evaluate mesh quality through five metrics: Accuracy, Completion, Chamfer-L1 distance, Normal Consistency and F-scores. For novel view synthesis, we follow standard PSNR, SSIM and LPIPS metrics for rendered images. 

\paragraph{Baselines.}
We compare our model against a series of baseline approaches on two tasks.
a) 3D open-vocabulary segmentation: Langsplat\cite{qin2024langsplat}, Gaussian Grouping\cite{gaussian_grouping} and OpenGaussian\cite{wu2024opengaussian}.
We deploy the scheme of OpenGaussian to render the selected objects. Specifically, we compute the similarity between 3D Gaussian semantic features and text features, then select Gaussians with high similarity and obtain the rendered object by the rasterizer. 
b) 3DGS-based indoor surface reconstruction: 2DGS\cite{Huang2DGS2024}, PGSR\cite{chen2024pgsr}, DN-Splatter\cite{turkulainen2024dnsplatter} and FDS\cite{chen2024fds}. For fair comparison among all methods, we take the LiDAR points of iPhone as the default input point cloud on both datasets, and re-ran their source codes released in GitHub in 5 scenes.
To align with the setting of DN-Splatter, we introduce the sensor depth to supervise $D_p$ by Mean Absolute Error loss.

\paragraph{Implementation details.}
Our code is built based on PGSR\cite{chen2024pgsr}. The densification strategy is adopted from AbsGS\cite{ye2024absgs}.
We train GLS for 30k iterations, consuming about 40 minutes on a single NVIDIA RTX 4090 GPU. 
For indoor surface reconstruction, we first generate rendered depth in each training view, followed by performing the TSDF Fusion\cite{newcombe2011kinectfusion,lorensen1998marching} to extract the mesh in the TSDF field. Subsequently, for 3D open-vocabulary segmentation, we first compute the similarity between 3D
Gaussian semantic features and text features, then filter Gaussians with high similarity to render and reconstruct the selected object.
More details can be seen in the supplementary materials. 

\subsection{Indoor Surface Reconstruction}
\paragraph{Quantitative Comparisons.}
\label{sec:exp_1}
For indoor surface reconstruction, we conduct comparisons on two real-world datasets, including MuSHRoom\cite{ren2024mushroom} and ScanNet++\cite{yeshwanth2023scannet++}.
We report the metric values in \tabref{tab:mush} and \tabref{tab:scanpp}. It can be seen that our method outperforms other 3DGS-based approaches\cite{Huang2DGS2024,chen2024pgsr} without sensor depth among all metrics. When adopting sensor depth as the prior information of scene scale, our model also achieves better performance of sharpness and smoothness than DN-Splatter\cite{turkulainen2024dnsplatter} according to the Accuracy and Normal Consistency metric. 
Compared with the \sArt method FDS\cite{chen2024fds}, our results have better smoothness, accuracy and training efficiency. 

\begin{figure*}[t]
\centering
\includegraphics[width=\textwidth]{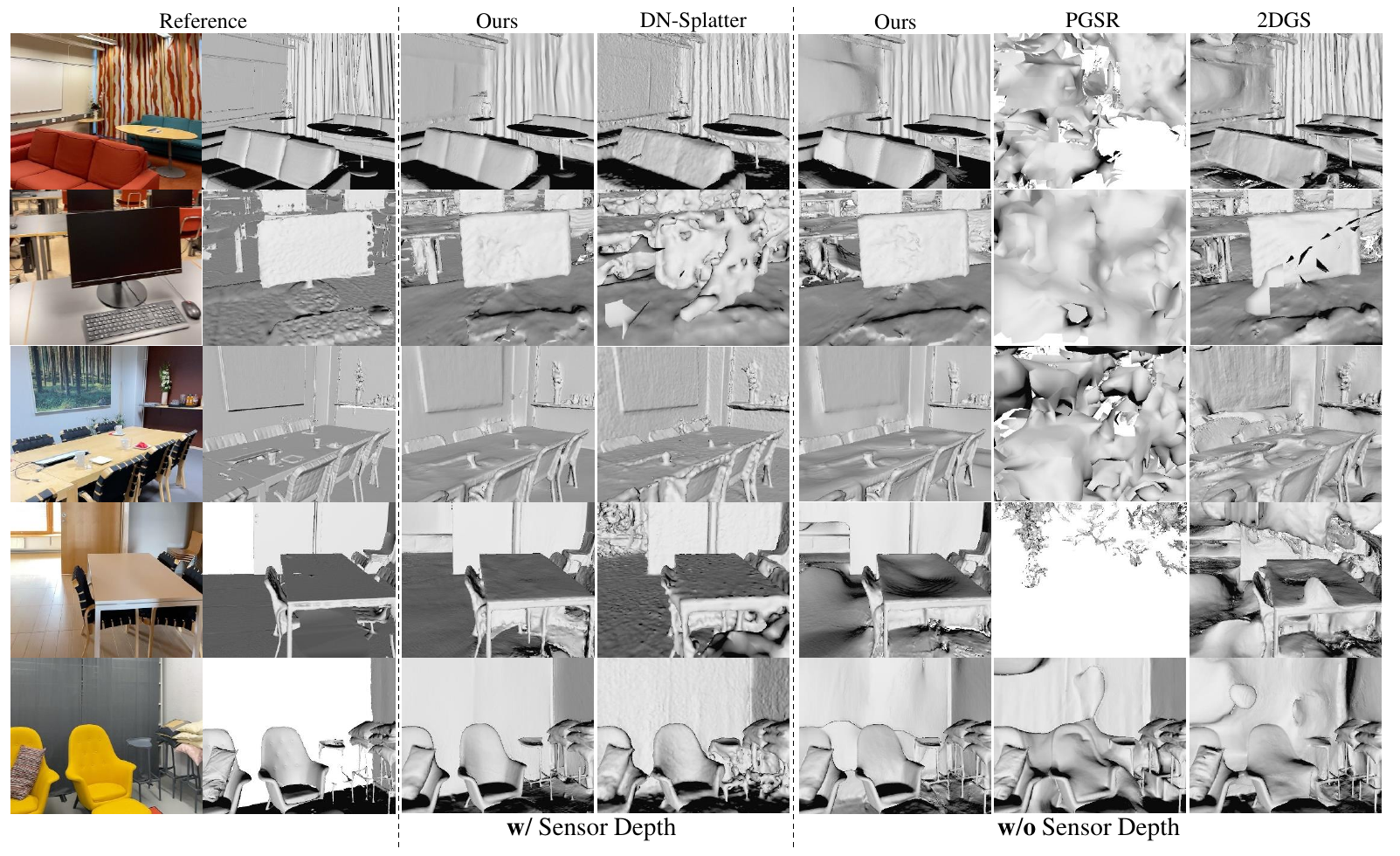}
\vspace*{-7mm}
\caption{Qualitative comparisons of indoor surface reconstruction on MuSHRoom dataset. PGSR generates unstable results by the default hyperparameters.}
\vspace*{-5mm}
\label{fig:exp_recon1}
\end{figure*}

\paragraph{Qualitative Comparisons.}
As shown in \figref{fig:exp_recon1}, there are surface reconstruction results produced by different methods on the MuSHRoom dataset. It can be observed that DN-Splatter\cite{turkulainen2024dnsplatter} generates severe noise on scene surfaces and even destroys object structures because of texture-less regions. On the contrary, our model reconstructs smooth and sharp scene surfaces and recovers more thin structures than ground-truth scene surfaces. This observation demonstrates the joint optimization of surface reconstruction and 3D open-vocabulary segmentation can significantly promote the reconstruction quality.
For 3DGS-based approaches without sensor depth as supervision, due to the complex camera motion and light conditions, PGSR presents unstable performance and fails in most scenes. When compared to 2DGS, our model handles shadow and high-light regions better and generates cleaner scene surfaces.  

\begin{figure*}[t]
\centering
\includegraphics[width=\textwidth]{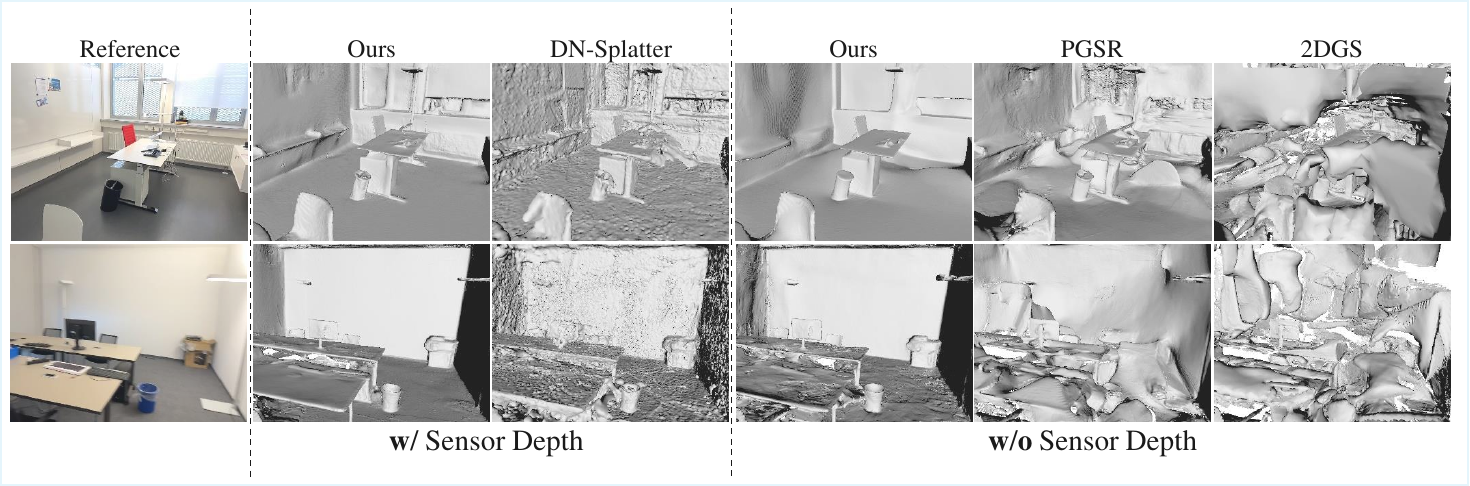}
\vspace*{-7mm}
\caption{Qualitative comparisons of indoor surface reconstruction on ScanNet++ dataset. The lighting conditions change violently in these scenes.}
\vspace*{-5mm}
\label{fig:exp_recon2}
\end{figure*}

Moreover, we further evaluate the generalization ability of all methods
on ScanNet++ dataset as \figref{fig:exp_recon2} shows. 
These scenes encode various lighting conditions, which makes reconstructing scene surfaces more challenging. Hence, more noises occur in DN-splatter results, while our results are still sharp and smooth. 
For 3DGS-based approaches without sensor depth as supervision, PGSR and 2DGS produce meaningless results while our results successfully tackle these scenes well.

\subsection{3D open-vocabulary segmentation}
\paragraph{Quantitative Comparisons.}
3D open-vocabulary segmentation drops predefined labels of objects and uses texts as prompts to segment target objects. 
We conduct comparisons of 3D open-vocabulary segmentation on the widely-used LERF-OVS dataset.
Following Gaussian grouping\cite{gaussian_grouping}, we employ mIoU and mBIoU as the evaluation metrics, which represent the segmentation quality of the selected object. The estimated metric scores are reported in \tabref{tab:lerf}. Our model outperforms \sArt
method OpenGaussian\cite{wu2024opengaussian} over $4\%$ on mIoU and achieves over $20\%$ gain when compared with Gaussian grouping\cite{gaussian_grouping} on mBIoU. This fact illustrates that the joint optimization of two tasks improves the smoothness and sharpness of scene surfaces then makes segmentation results smoother and sharper.
\label{sec:exp_2}
\begin{table*}[t]
\footnotesize
\centering
\caption{Quantitative results of 3D open-vocabulary segmentation on LERF-OVS dataset. We follow the metrics of LangSplat\cite{qin2024langsplat} and OpenGaussian\cite{wu2024opengaussian} from the paper of OpenGaussian. The best metrics are \textbf{highlighted}.}
\vspace*{-3mm}
\resizebox{\linewidth}{!}{\begin{tabular}{c|ccccc|ccccc|c}
\cline{1-12}
\multirow{2}{*}{Methods} & \multicolumn{5}{c}{mIoU$\uparrow$} & \multicolumn{5}{|c}{mBIoU$\uparrow$} & \multirow{2}{*}{Time}  \\
 & figurines & ramen & teatime & waldo\_kitchen & \textbf{Mean} & figurines & ramen & teatime & waldo\_kitchen & \textbf{Mean} &\\
\cline{1-12}
LangSplat\cite{qin2024langsplat} & 10.16 & 7.92 & 11.38 & 9.18 & 9.66 & - & - & - & - & - & 2.2h\\
Gaussian grouping\cite{gaussian_grouping} & 15.53 & 17.49 & 22.27  & 26.51 & 20.45 & 13.71 & 13.86 & 19.10 & 18.85 & 16.38 & 0.7h\\
OpenGaussian\cite{wu2024opengaussian} & 39.29 & 31.01 & \textbf{60.44} & 
22.70 & 38.36 & - & - & - & - & - & 1.0h\\
Ours & \textbf{49.73} & \textbf{31.21} & 58.01 & \textbf{32.15} & \textbf{42.78} & \textbf{47.97} & \textbf{27.90} & \textbf{52.96} & \textbf{23.93} & \textbf{38.19} & 0.7h\\
\cline{1-12}
\end{tabular}}
\label{tab:lerf}
\end{table*}

\begin{figure}[t]
\centering
\includegraphics[width=\linewidth]{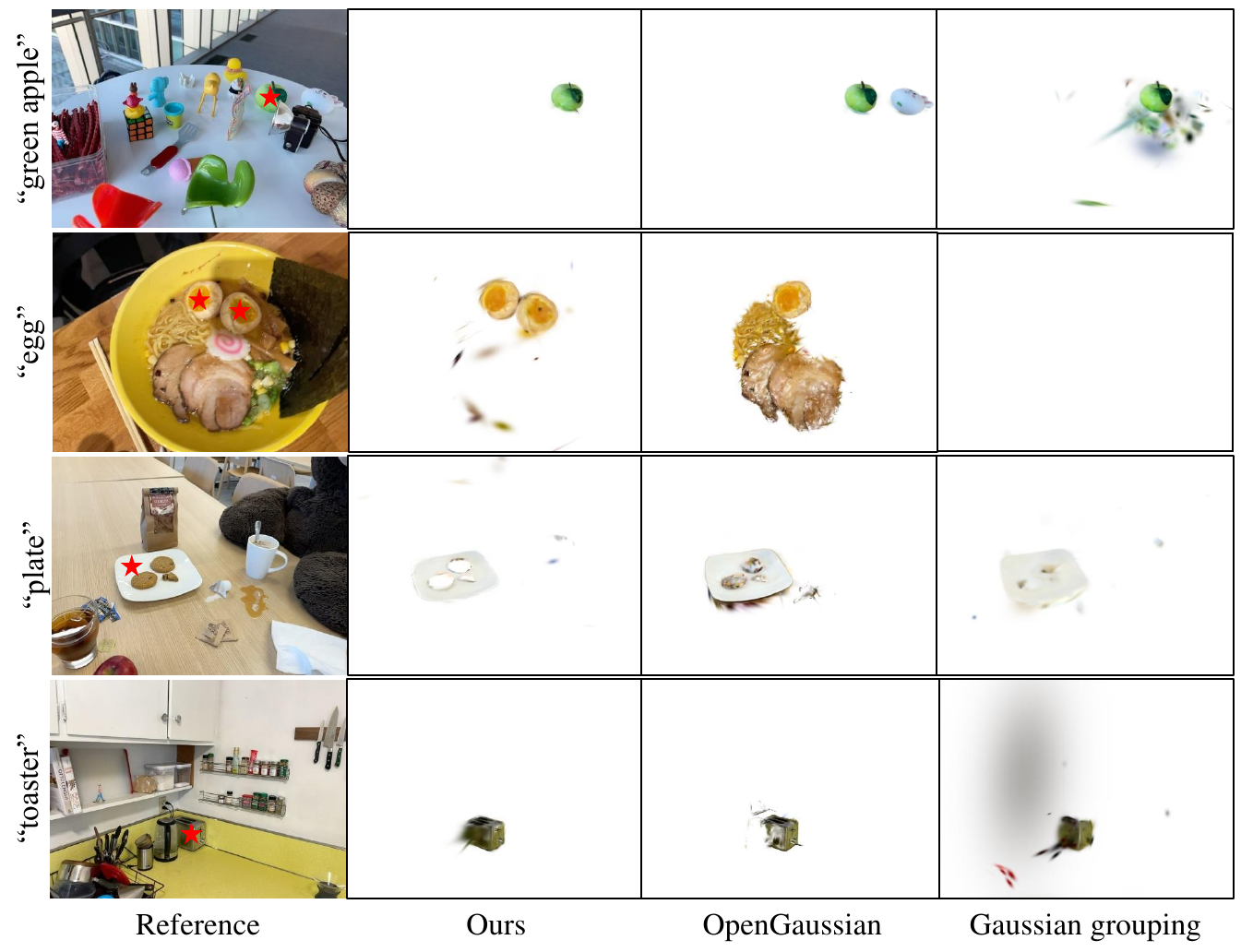}
\vspace*{-7mm}
\caption{Qualitative comparisons of 3D open-vocabulary segmentation on LERF-OVS dataset, between \sArt methods (OpenGaussian\cite{wu2024opengaussian} and Gaussian grouping\cite{gaussian_grouping}) and our model.}
\vspace*{-3mm}
\label{fig:exp_seg}
\end{figure}

\paragraph{Qualitative Comparisons.}
\figref{fig:exp_seg} shows some 3D open-vocabulary segmentation results of different methods. Without the optimization of surface reconstruction, OpenGaussian\cite{wu2024opengaussian} easily treats other objects as targets and Gaussian
grouping\cite{gaussian_grouping} exhibits noisy results. Our model successfully identifies the 3D Gaussians relevant to the query text and generates object selection with a clearer boundary. 

\subsection{Ablation study}
We conduct sufficient ablation experiments to study the effect of different regularization terms, by disabling each one and enabling others. The results of indoor surface reconstruction and 3D open-vocabulary segmentation are reported in \tabref{tab:ab_1} and
\tabref{tab:ab_2} respectively.

\label{sec:exp_ab}
\begin{table}[t]
\footnotesize
\centering
\caption{Ablation study of indoor surface reconstruction without the sensor depth. The best metrics are \textbf{highlighted}. The \fcolorbox{red}{white}{red box} means that the loss function is designed for 3D open-vocabulary segmentation.}
\vspace*{-3mm}
\resizebox{\linewidth}{!}{\begin{tabular}{c|ccc}
\cline{1-4}
 Settings & Accuracy$\downarrow$ & Normal Consistency$\uparrow$ & F-score$\uparrow$ \\
\cline{1-4}
No $\mathcal{L}_n$ & 0.1850 & 0.6893 & 0.2322 \\
No $\mathcal{L}_d$ & 0.1077 & 0.8023 & 0.4440 \\
No $\mathcal{L}_s$ & 0.1303  & 0.7910 & 0.4356 \\
\fcolorbox{red}{white}{No $\mathcal{L}_{clip}$} & 0.0966 & 0.8393 &  0.4614\\
\fcolorbox{red}{white}{No $\mathcal{L}_m$} & 0.0951 & 0.8425 &  0.4671\\
\cline{1-4}
All & \textbf{0.0814} & \textbf{0.8474} & \textbf{0.5127}\\
\cline{1-4}
\end{tabular}}
\label{tab:ab_1}
\vspace*{-3mm}
\end{table}

\begin{table}[t]
\footnotesize
\centering
\caption{Ablation study of 3D open-vocabulary segmentation. The best metrics are \textbf{highlighted}. The \fcolorbox{red}{white}{red box} means that the loss function is designed for indoor surface reconstruction.}
\vspace*{-3mm}
\resizebox{0.5\linewidth}{!}{\begin{tabular}{c|c|c}
\cline{1-3}
 Settings & mIoU$\uparrow$ & mBIoU$\uparrow$ \\
\cline{1-3}
\fcolorbox{red}{white}{No $\mathcal{L}_n$} & 32.90 & 27.61\\
\fcolorbox{red}{white}{No $\mathcal{L}_d$} & 31.81 & 27.37\\
\fcolorbox{red}{white}{No $\mathcal{L}_s$} & 33.98 & 29.08\\
No $\mathcal{L}_{clip}$ & 29.14 & 25.01\\
No $\mathcal{L}_m$ & 28.71 & 22.62\\
\cline{1-3}
All & \textbf{42.78} & \textbf{38.19}\\
\cline{1-3}
\end{tabular}}
\label{tab:ab_2}
\end{table}

\paragraph{Effect of $\mathcal{L}_n$.}
The Normal prior supplies the regions with smooth surfaces, where shadow and highlight exist in an indoor scene. For indoor surface reconstruction, $\mathcal{L}_n$ remarkably improves surface smoothness in terms of the  Normal Consistency metric. For 3D open-vocabulary segmentation, $\mathcal{L}_n$ helps the full model to produce accurate boundary of segmentation results by the joint optimization according to the metric mBIoU.  

\paragraph{Effect of $\mathcal{L}_d$.}
\figref{fig:a2} illustrates the visual effects of $\mathcal{L}_d$ for indoor surface reconstruction. For 3D open-vocabulary segmentation, the performance of our model degrades when disabling $\mathcal{L}_d$. Although $\mathcal{L}_d$ is designed for refining the unbiased depth, it also strengthens the quality of segmentation results through joint optimization.

\paragraph{Effect of $\mathcal{L}_s$.}
To determine whether $\mathcal{L}_s$ is necessary for two tasks, we disable it from our full model. $\mathcal{L}_s$ enhances the performance of our model among all metrics, by connecting the surface normal estimated from the unbiased depth and CLIP features.

\paragraph{Effect of $\mathcal{L}_{clip}$.}
Then we disable $\mathcal{L}_{clip}$, to verify it is a significant regularization term for two tasks. We introduce CLIP features to strengthen the representation ability of Gaussian semantic features, and which plays the most important role in 3D open-vocabulary segmentation. It also significantly maintains the smoothness and sharpness of reconstructed indoor surfaces via joint optimization. 

\paragraph{Effect of $\mathcal{L}_m$.}
We further explore the effect of $\mathcal{L}_m$ by dropping the segmentation results of DEVA as supervision. As \tabref{tab:ab_2} reports, $\mathcal{L}_m$ faithfully increases the sharpness of  rexonstruction and segmentation results. 

\subsection{Novel view synthesis}
Finally, we evaluate the performance of our model in the task of novel view synthesis. The results are presented in \tabref{tab:nvs}. Our model achieves comparable performance with DN-Splatter, and better performance against other geometry-aware approaches in the field of indoor surface reconstruction. We also compare our method with OpenGaussian in the field of 3D open-vocabulary segmentation, our method renders more accurate novel views than it.

\begin{table}[t]
\footnotesize
\centering
\caption{Quantitative results of novel view synthesis on ScanNet++ 
 and LERF-OVS dataset. The best metrics are \textbf{highlighted}.}
\vspace*{-3mm}
\resizebox{\linewidth}{!}{\begin{tabular}{c|c|ccc}
\cline{1-5}
Dataset & Methods & PSNR$\uparrow$ & SSIM$\uparrow$ & LPIPS$\downarrow$ \\
\cline{1-5}
\multirow{4}{*}{ScanNet++} & DN-Splatter\cite{turkulainen2024dnsplatter} & \textbf{20.73} & 0.8476 & 0.1996 \\
& 2DGS\cite{Huang2DGS2024} & 15.60 & 0.7791 & 0.2581 \\
& PGSR\cite{chen2024pgsr} & 19.47 & 0.8275 & 0.2060 \\
& Ours & 20.69 & \textbf{0.8496} & \textbf{0.1709}\\
\cline{1-5}
\multirow{2}{*}{LERF-OVS} & OpenGaussian\cite{wu2024opengaussian} & 23.78 & 0.8508 & \textbf{0.2295} \\
& Ours & \textbf{24.13} & \textbf{0.8537} & 0.2427\\
\cline{1-5}
\end{tabular}}
\vspace*{-3mm}
\label{tab:nvs}
\end{table}

\section{Conclusion}
\label{sec:con}
In this work, we present GLS, a novel 3DGS-based framework that effectively combines indoor surface reconstruction and 3D open-vocabulary segmentation. We propose leveraging 2D geometric and semantic cues to optimize the performance of 3DGS on two tasks jointly. We design two novel regularization terms to enhance the sharpness and smoothness of the scene surface, and then improve the segmentation quality. Comprehensive experiments on both 3D open-vocabulary segmentation and indoor surface reconstruction tasks illustrate that GLS outperforms \sArt methods quantitatively and qualitatively. Besides, the ablation study explores the effectiveness of each regularization term on two tasks.
\paragraph{Limitation.}
The proposed method follows the natural limitation of TSDF fusion, whose completeness relies on the number of captured views. Our model generates empty geometry of selected objects in unseen views of the scene. Introducing image-to-3D models to supply unseen information can help solve this issue.
{
    \small
    \bibliographystyle{ieeenat_fullname}
    \bibliography{main}
}

\appendix

\newcounter{question}
\setcounter{question}{0}

\newcommand{\question}[1]{\item[Q\refstepcounter{question}\thequestion.] \textit{#1}}
\newcommand{\answer}[1]{\item[A\thequestion.] #1}
\DeclareFixedFont{\ttb}{T1}{txtt}{bx}{n}{8} 
\DeclareFixedFont{\ttm}{T1}{txtt}{m}{n}{8}  

\definecolor{deepblue}{rgb}{0,0,0.5}
\definecolor{deepred}{rgb}{0.6,0,0}
\definecolor{deepgreen}{rgb}{0,0.5,0}
\definecolor{myRed}{rgb}{0.8,0.2,0.2}
\definecolor{myBlue}{rgb}{0.2,0.2,0.8}
\def\sArt{state-of-the-art~}



Unlike recent methods of indoor surface reconstruction~\cite{ren2024agsmesh,turkulainen2024dnsplatter,chen2024fds} and 3D open-vocabulary segmentation~\cite{wu2024opengaussian} only concentrate on optimizing one task, our goal is to strengthen two tasks via our joint optimization scheme. Here, we provide more details of our method and experiments. Specifically, we provide analysis through a Q\&A section to answer four important questions (\secref{Sec:qa}), additional algorithm details (\secref{Sec:aad}), additional details of experiments (\secref{Sec:aed}), additional results (\secref{Sec:ar}) and the future work (\secref{Sec:fw}).

\begin{figure*}[t]
\centering
\includegraphics[width=\linewidth]{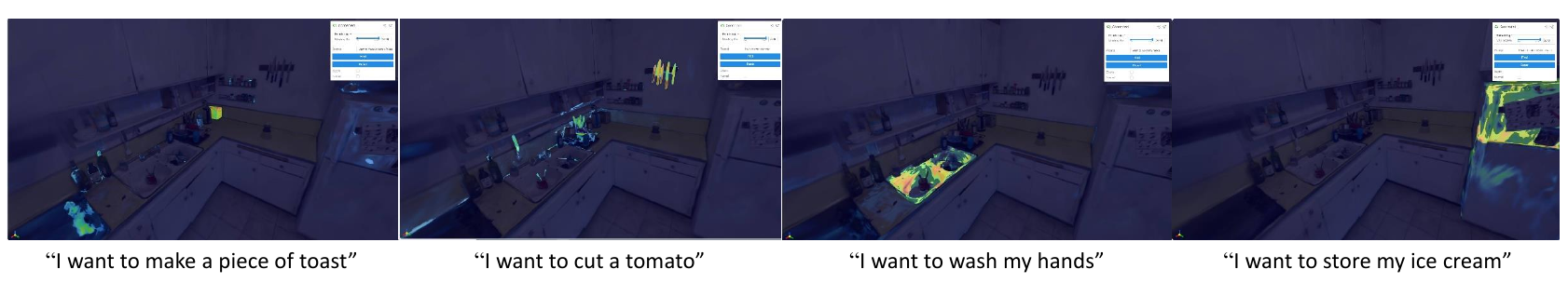}
\caption{A demo of \textbf{GLS + nerfview + GPT-4V}. Our tool can find the object that solves the user's request (bottom) and extract the scene geometry.}
\label{fig:q4}
\end{figure*}

\begin{figure*}[t]
\centering
\includegraphics[width=\linewidth]{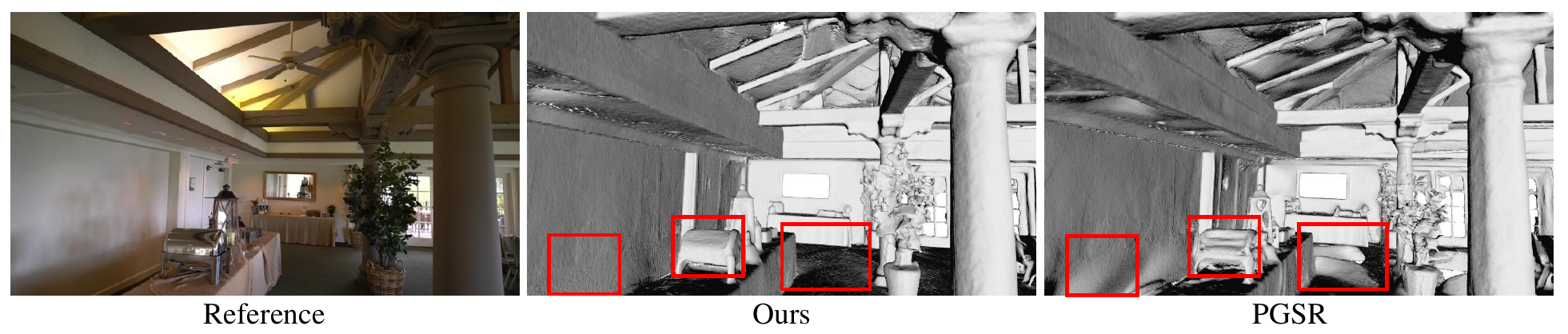}
\caption{Qualitative results in a large-scale indoor scene. Our method successfully handles the high-light and dark regions.}
\label{fig:q2}
\end{figure*}

\begin{figure*}[t]
\centering
\includegraphics[width=\linewidth]{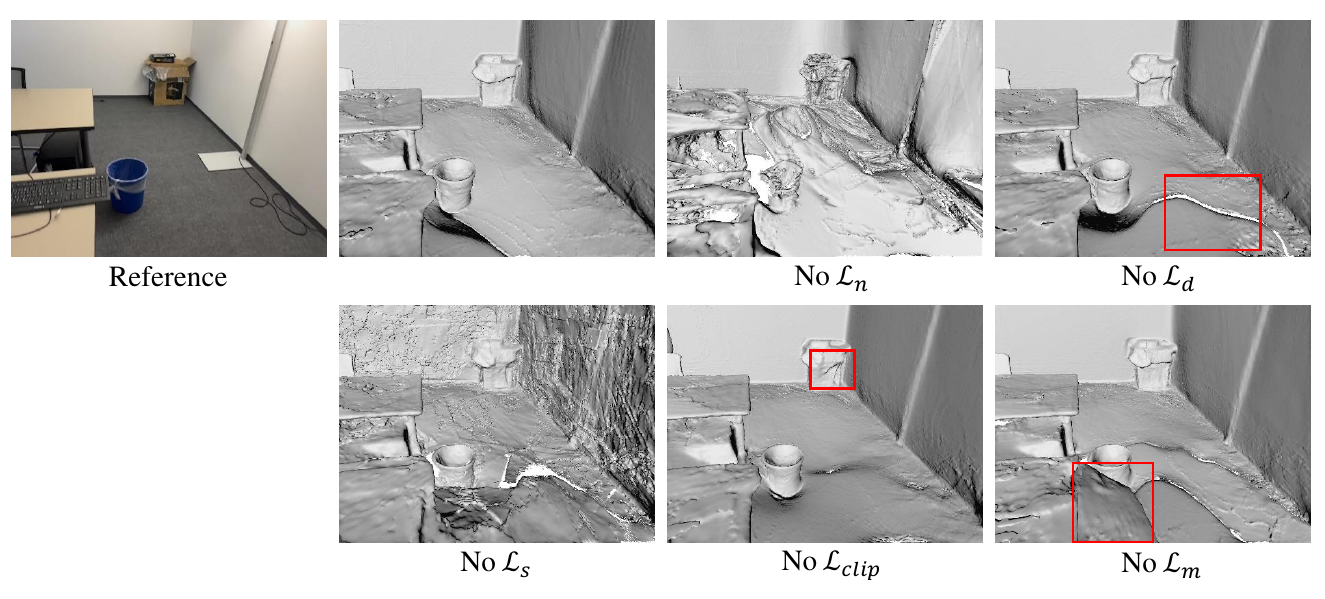}
\caption{Qualitative results of the ablation study on `b20a261fdf' of ScanNet++ dataset. The performance of GLS degenerates reasonably when we disable a regularization term and enable others.}
\label{fig:q1}
\end{figure*}

\section{Q\&A}\label{Sec:qa}
\begin{figure}[t]
\centering
\includegraphics[width=\linewidth]{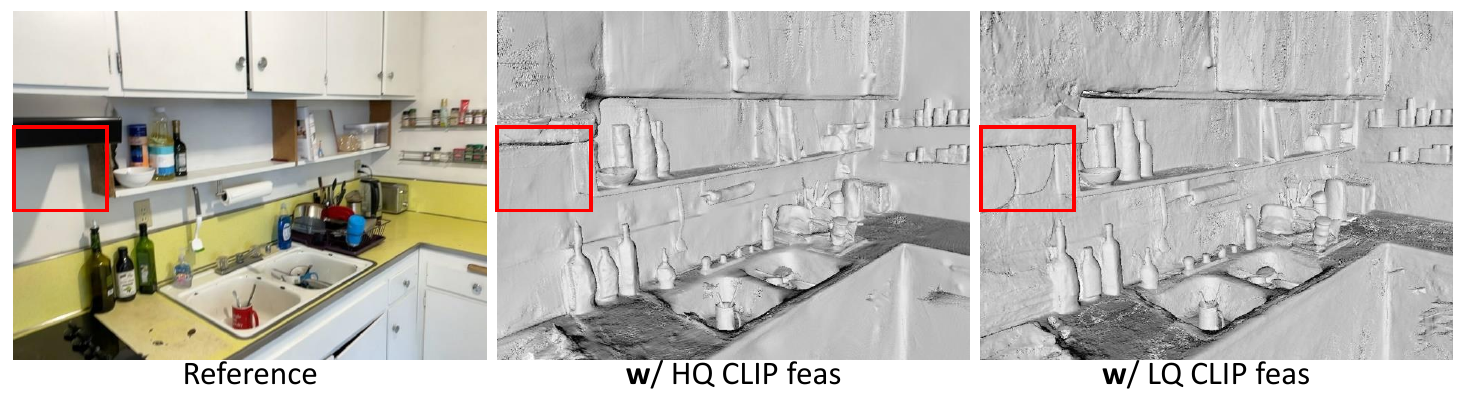}
\caption{Effect of different quality of CLIP features.}
\label{fig:weak_sem2}
\end{figure}

\begin{figure}[t]
\centering
\includegraphics[width=\linewidth]{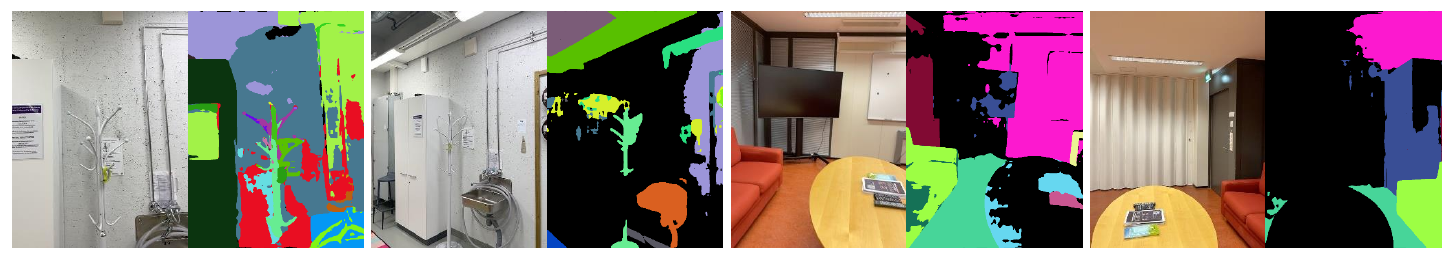}
\caption{Example of the view-consistent segmentation results of DEVA [11]. We directly use these results as
the supervision of the rendered mask.}
\label{fig:weak_sem1}
\end{figure}

\begin{figure}[t]
\centering
\includegraphics[width=\linewidth]{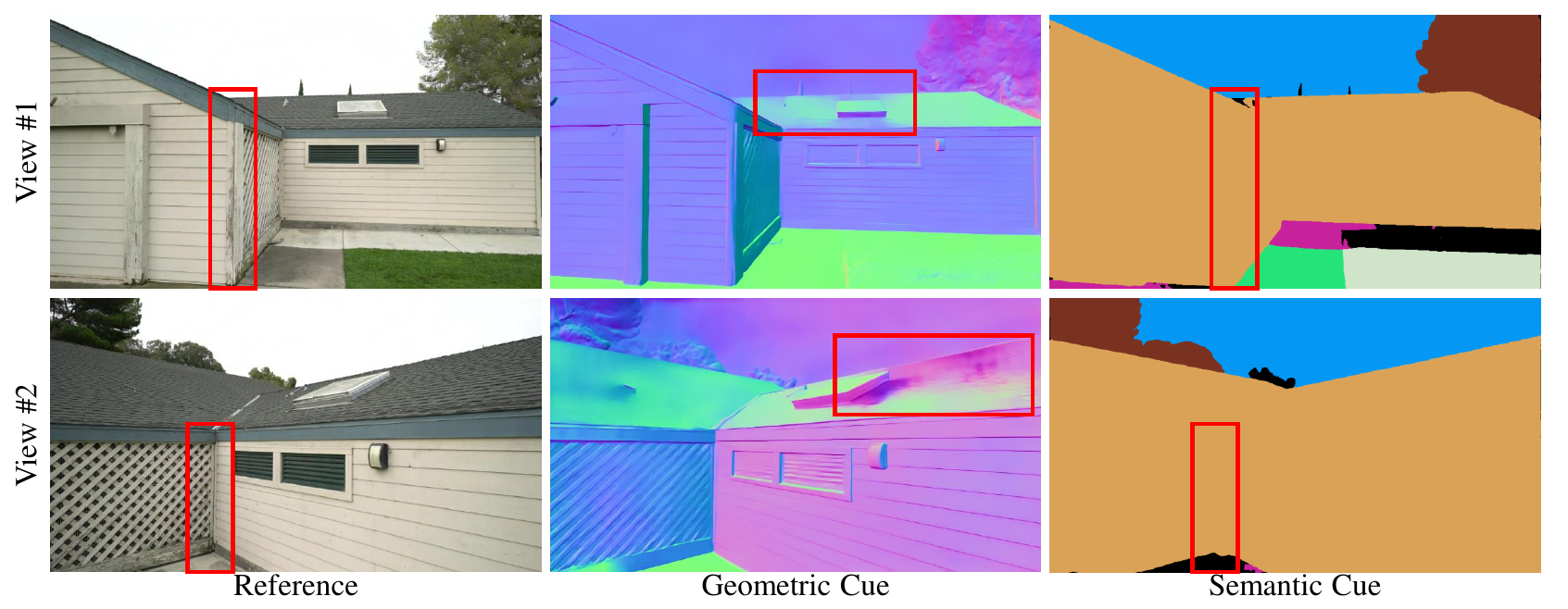}
\caption{Limitation of geometric and semantic cues in outdoor scenes. The geometric cues predicted from the pretrained monocular normal model, suffer from inconsistent noises across different views. The semantic cues encode large ambiguity of object structures.}
\label{fig:geo_sem}
\end{figure}

\begin{itemize}
\question{\textbf{Please explain the benefits of the proposed joint optimization.}}
\answer{We have analyzed the benefits of the proposed joint optimization of indoor surface reconstruction and 3D open-vocabulary segmentation from Line\_037 to Line\_056 of the manuscript. Theoretically, the optimization goals of the two tasks are the \textbf{smoothness} and \textbf{sharpness} of generated results (including the surface mesh and the segmentation mask). Then we can optimize two tasks simultaneously through the gradient descent strategy.  
The marked metrics in Table~4 and Table~5 of the manuscript demonstrate the effectiveness of the proposed joint optimization in two fields. The qualitative results of the ablation study are shown in ~\figref{fig:q1}. The performance of GLS degenerates reasonably when we disable a regularization term and enable others.}

\question{\textbf{Why does GLS only focus on indoor scenes?}}
\answer{
Considering the unique challenges of indoor scenes and the limitation of geometric and semantic cues in outdoor scenes, there are two main reasons as follows:
\begin{itemize}[leftmargin=*] 
\item[i.] For surface reconstruction, we follow DN-Splatter~\cite{turkulainen2024dnsplatter} and FDS~\cite{chen2024fds}, then focus on achieving high-fidelity reconstruction in indoor scenes. Unlike outdoor scenes (e.g., the Tanks and Temples dataset~\cite{Knapitsch2017}), indoor scenes suffer from more challenging conditions. 
On the one hand, indoor scenes are always captured by free camera trajectories, which causes the imbalanced allocation of spatial representation capacity. 
On the other hand, there are a lot of texture-less regions (such as white walls and floors) in indoor scenes. 
These issues cause extreme ambiguity of multi-view reconstruction and large performance degeneration of the \sArt method PGSR~\cite{chen2024pgsr}. However, GLS tackles these issues well because of the robust joint optimization.

\item[ii.] As~\figref{fig:geo_sem} shows, geometric cues predicted from the pre-trained monocular normal model, suffer from inconsistent noises across different views in outdoor scenes. The semantic cues only consist of a few instances and encode large ambiguity of object structures. These issues are harmful to the joint optimization of GLS. 

Hence, we focus on the surface indoor scenes where the quality of geometric cues is more stable and semantic distribution is highly predictable.
To illustrate the generalization ability of GLS, 
We also evaluate the performance of our model in a large-scale indoor scene (`Meetingroom') of the Tanks and Temples dataset, the result is presented in ~\figref{fig:q2}. Our method successfully handles the high-light and dark regions and reconstructs a smoother scene surface than PGSR~\cite{chen2024pgsr}.
\end{itemize}
}
\question{\textbf{How about the robustness of GLS under weak semantic priors?}}
\answer{Most view-consistent segmentation priors we used on the MuSHRoom dataset are noisy or incomplete, such as ~\figref{fig:weak_sem1} shows.
It demonstrates that our method is certainly robust.
We also evaluate the reconstruction result under different quality of CLIP features, by disabling the denoising procedure of pre-processing to obtain the low-quality CLIP features. The result is shown in ~\figref{fig:weak_sem2}. 
High-quality CLIP features reduce shadow interference. This result also illustrates the effectiveness of the proposed joint optimization.}

\question{\textbf{How about applications of GLS?}}
\answer{
Thanks to the high-quality surface reconstruction and segmentation results of GLS, we develop several tools to explore applications of GLS:
\begin{itemize}[leftmargin=*] 
\item[i.] \textbf{GLS + Blender}: With the help of Blender~\cite{blender}, we can achieve the scene-editing effects, including moving the target object and adding objects.
\item[ii.] \textbf{GLS + nerfview}: We develop an interactive tool based on nerfview~\cite{nerfview}, to show the scene geometry (depth and normal) and open-vocabulary attention in free views.
\item[iii.] \textbf{GLS + nerfview + GPT-4V}: As ~\figref{fig:q4} shows, we also add the GPT-4V~\cite{achiam2023gpt} to `ii' to achieve intelligent interaction. Given a user's request, our tool can find the object that solves the request and extract the scene geometry used to help the user plan the path to the target object. 
We believe this tool has potential value in the field of embodied intelligence.
\end{itemize}
Our supplementary videos consist of two application demos (`application\_i.mp4' and `application\_ii.mp4'). 
We will release all tools when this paper is accepted.
}

\end{itemize}

\section{Additional Algorithm Details}\label{Sec:aad}
\myPara{About $N_d$.}
We follow the manner of PGSR~\cite{chen2024pgsr} to estimate $N_d$.
Given a pixel point p and its four neighboring
pixels, we unproject these 2D points into 3D points $\{P_i|i=0,...,4\}$ by $D_p$, then calculate the local normal $N_d$ of p via:
\begin{equation}
\setlength{\abovedisplayskip}{3pt}
\setlength{\belowdisplayskip}{3pt}
N_d(p) = \frac{(P_1-P_0)\times(P_3-P_2)}{|(P_1-P_0)\times(P_3-P_2)|}
\end{equation}

\begin{figure}[t]
\centering
\includegraphics[width=\linewidth]{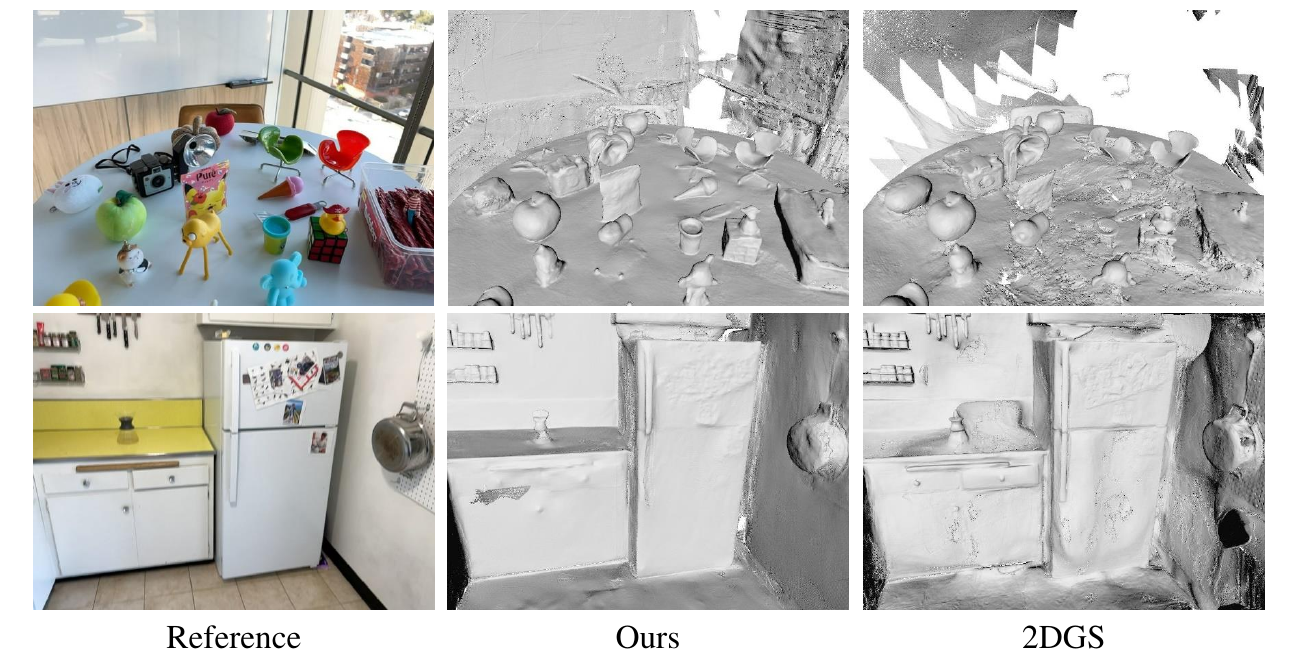}
\caption{Results of indoor surface reconstruction on LERF-OVS dataset. Our model reconstructs sharper and smoother indoor surfaces than 2DGS~\cite{Huang2DGS2024}.}
\label{fig:exp_recon3}
\end{figure}

\myPara{About $\mathcal{L}_n$.}
Inspired by the 2DGS~\cite{Huang2DGS2024} and considering semi-transparent surfels, we adaptively give high weights to actual surfaces in $\mathcal{L}_n$ by $A$.

\myPara{About $\mathcal{L}_s$.}
$\mathcal{L}_s$ is a powerful regularization term for $N_d$ and easily causes over-smoothing effect for small objects.
In practice, we adaptively select big objects that occupy top-$3$ areas in a view by $M_o$. Then we adopt $\mathcal{L}_s$ to constrain them.

\myPara{About $D_r$ and $\mathcal{L}_d$.}
Essentially, $D_r$ is a refined depth based on the probability rather than an actual depth. It is built on the error analysis between the rendered normal and the ideal normal. We only consider the normal prior~\cite{bae2024dsine} as the reference normal, because its quality is unstable as mentioned by VCR-GauS~\cite{chen2024vcr}. Hence, the ideal normal cannot be acquired and the error analysis is necessary. 
We adopt a total of six loss functions during training, to balance the value of each loss function and avoid the abrupt value of $D_r$, we propose the function $y = 1-e^{(-|x|)}$ to generate $\mathcal{L}_d$.

\section{Additional Experimental Details} \label{Sec:aed}
\tabref{tab:data} presents metrics of datasets used in our experiments. We list additional optimization hyperparameters below:
\begin{python}
def __init__(self, parser):
    self.iterations = 30_000
    self.position_lr_init = 0.00016
    self.position_lr_final = 0.0000016
    self.position_lr_delay_mult = 0.01
    self.position_lr_max_steps = 30_000
    self.feature_lr = 0.0025
    self.opacity_lr = 0.05
    self.scaling_lr = 0.005
    self.rotation_lr = 0.001
    self.percent_dense = 0.01
    self.lambda_dssim = 0.2
    self.densification_interval = 100
    self.opacity_reset_interval = 3000
    self.densify_from_iter = 500
    self.densify_until_iter = 15_000
    self.densify_grad_threshold = 0.0006
    self.opacity_cull_threshold = 0.05
    self.densify_abs_grad_threshold = 0.0008
    self.abs_split_radii2D_threshold = 20
    self.max_abs_split_points = 50_000
    self.max_all_points = 6000_000
\end{python}

For Open-vocabulary Segmentation, we follow LangSplat~\cite{qin2024langsplat} to decrease the last dimension of original CLIP features from $512$ to $16$ by the encoder part of an encoder-decoder network. Then we use the decoder part of the same network to increase the the last dimension of Gaussian semantic features from $16$ to $512$, to compute the relevancy between them and the original CLIP features.
The extraction scheme for SAM masks and CLIP features is also aligned with LangSplat~\cite{qin2024langsplat}, while we follow OpenGaussian~\cite{wu2024opengaussian} only to extract the large layer of SAM masks.
The learning rate of the MLP layer is $0.00005$.

For indoor surface reconstruction, we use the iPhone sequences
with COLMAP registered poses on both MuSHRoom~\cite{ren2024mushroom} and ScanNet++~\cite{yeshwanth2023scannet++} datasets. Besides, we adopt
the densification strategy of AbsGS~\cite{ye2024absgs}. 
We disable the exposure compensation strategy of PGSR to maintain fair comparison. 

\section{Additional results} \label{Sec:ar}
\begin{figure*}[t]
\centering
\includegraphics[width=0.9\linewidth]{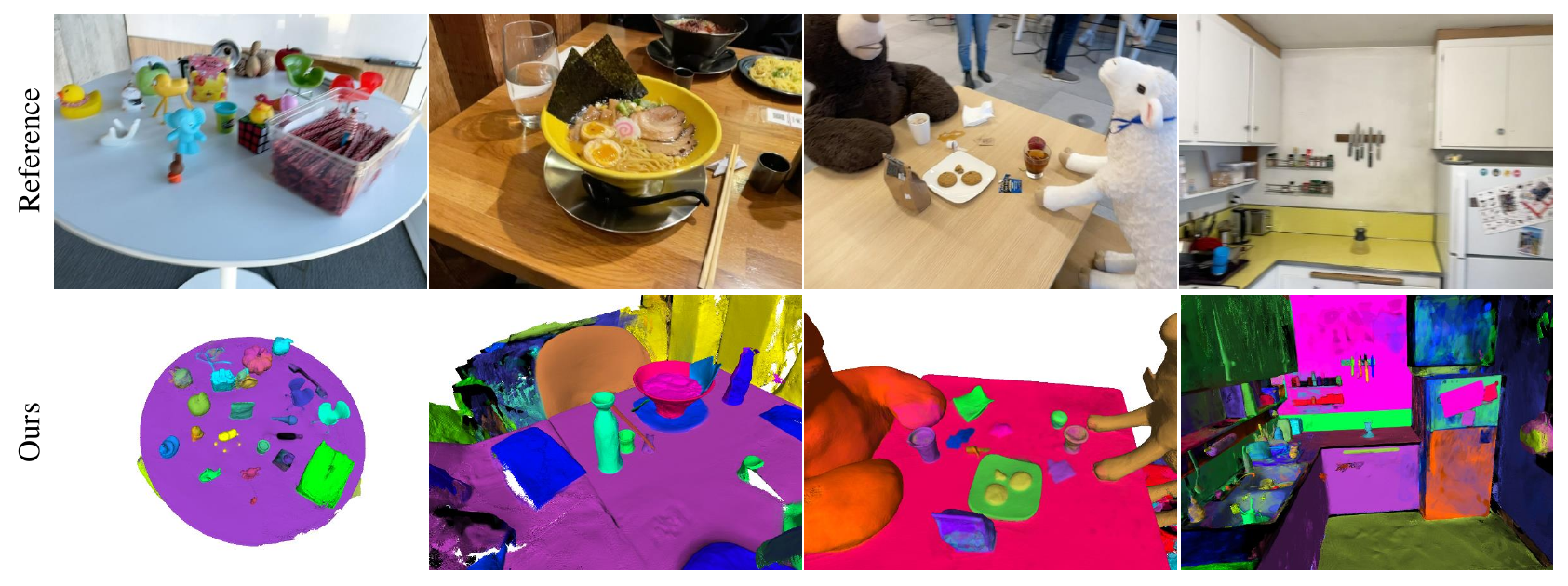}
\caption{Semantic mesh results of GLS on LERF-OVS~\cite{kerr2023lerf}.}
\label{fig:supp1}
\end{figure*}

\begin{figure*}[t]
\centering
\includegraphics[width=\linewidth]{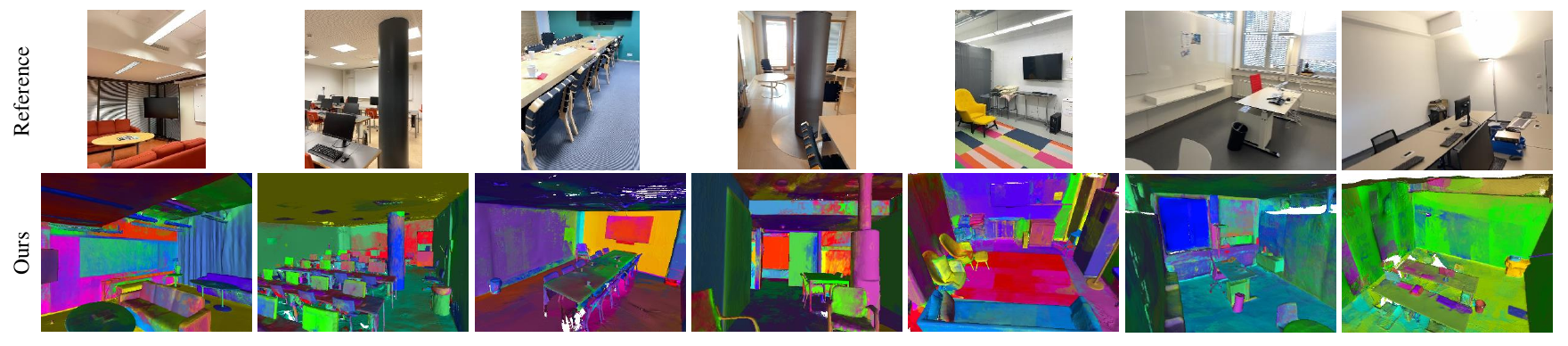}
\caption{Semantic mesh results of GLS on MuSHRoom~\cite{ren2024mushroom} and ScanNet++~\cite{yeshwanth2023scannet++}.}
\label{fig:supp2}
\end{figure*}

\begin{figure*}[t]
\centering
\includegraphics[width=\linewidth]{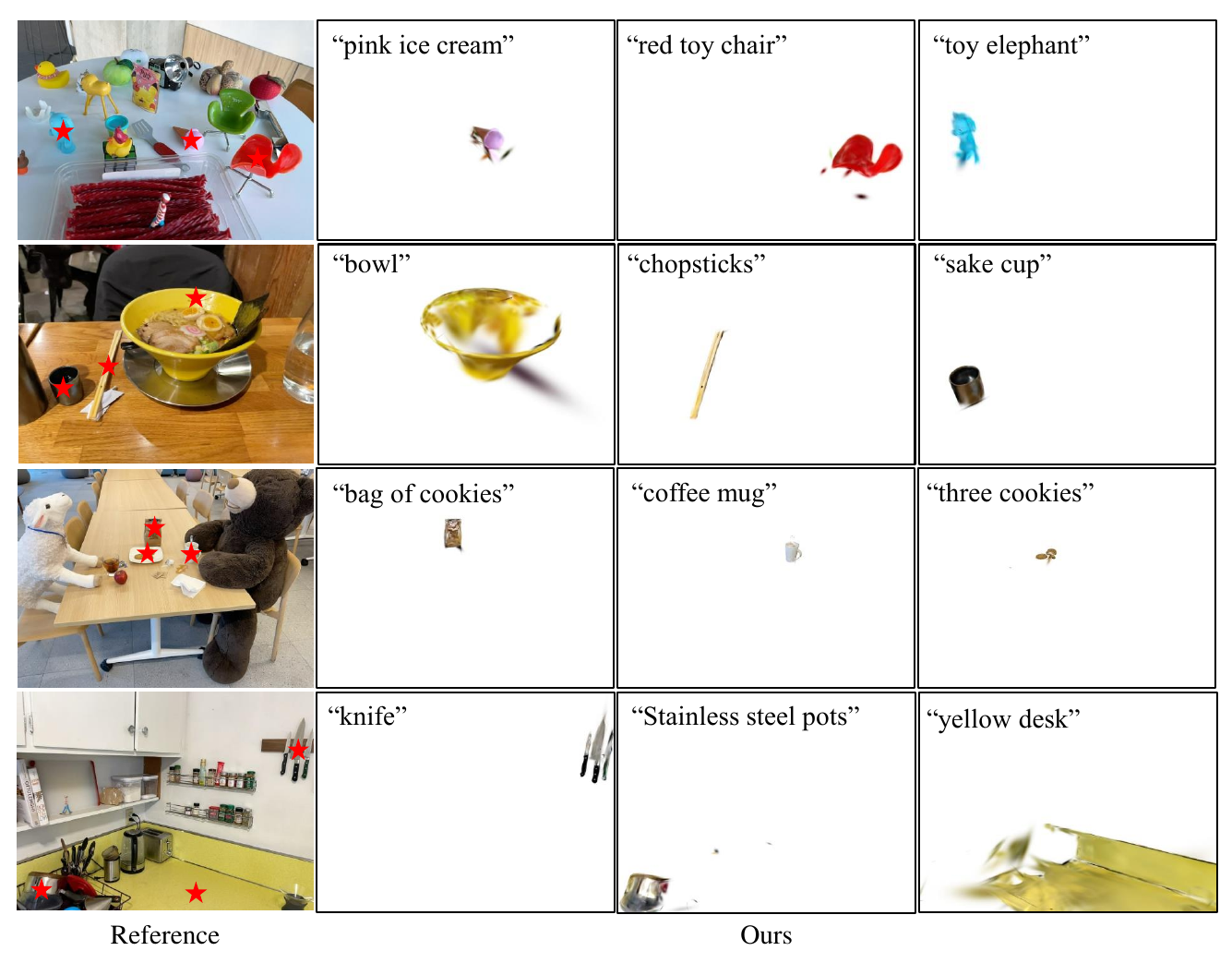}
\caption{More open-vocabulary segmentation results of GLS on LERF-OVS~\cite{kerr2023lerf}.}
\label{fig:supp3}
\end{figure*}

Per-scene quantitative results of GLS on the MuSHRoom and ScanNet++ datasets are reported in \tabref{tab:scanpp}.
As ~\figref{fig:supp1} and  ~\figref{fig:supp2} show, GLS also can attach the semantic information to the resconstructed mesh, by replacing the color image with the encoded semantic mask during TSDF fusion~\cite{newcombe2011kinectfusion}. Please note that our model is trained without any manual semantic annotations.
\figref{fig:supp3} shows more open-vocabulary segmentation results of GLS. Our model can accurately segment target objects selected by the corresponding text.

\figref{fig:exp_recon3} shows additional reconstruction results of our model and 2DGS on LERF-OVS dataset. Our model not only produces smooth and sharp object surfaces, but also successfully reconstructs the surfaces of highly- reflective whiteboard and transparent glass.

\section{Future Work}\label{Sec:fw}
In the future, we have two plans:
\begin{itemize}[leftmargin=*]
\item[i.] To address the limitation of TSDF Fusion, we can first segment object mesh based on GLS to supply the object scale and orientation, then leverage the image-to-3D models to obtain the whole object body from the segmented object appearance. 
\item[ii.] To extend GLS to outdoor scenes, we can follow VCR-GauS~\cite{chen2024vcr} to estimate the confidence of geometric and semantic cues to solve their issues. 
\end{itemize}

\begin{table*}[t]
\footnotesize
\centering
\caption{Per-scene quantitative results of GLS on the MuSHRoom and ScanNet++ datasets.}
\resizebox{\linewidth}{!}{\begin{tabular}{c|c|ccccc}
\cline{1-7}
 & Scenes & Accuracy$~\downarrow$ & Completion$~\downarrow$ & Chamfer$-L_1$$~\downarrow$ & Normal Consistency$~\uparrow$ & F-score$~\uparrow$ \\
\cline{1-7}
\multirow{7}{*}{\textbf{w}/ Sensor Depth} & coffee\_room  & 0.0231 & 0.0275 & 0.0253 & 0.8702 & 0.9020 \\
& computer & 0.0394 & 0.0277 & 0.0336 & 0.8756 & 0.8445 \\
& honka & 0.0264 & 0.0284 & 0.0274 & 0.8742 & 0.9090 \\
& kokko & 0.0305 & 0.0272 & 0.0444 & 0.9064 & 0.8623 \\
& vr\_room & 0.0244 & 0.0237 & 0.0241 & 0.8885 & 0.8802 \\
& 8b5caf3398 & 0.0936 & 0.0241 & 0.0588 & 0.8657 & 0.8741 \\
& b20a261fdf & 0.0335 & 0.0270 & 0.0303 & 0.9219 & 0.8841 \\
\cline{1-7}
\multirow{7}{*}{\textbf{w}/\textbf{o} Sensor Depth} & coffee\_room & 0.0684 & 0.0669 & 0.0676 & 0.7992 & 0.6103 \\
& computer & 0.0963 & 0.0874 & 0.0918 & 0.7708 & 0.4739 \\
& honka & 0.0810 & 0.0750 & 0.0780 & 0.7899 & 0.5503 \\
& kokko & 0.1102 & 0.1041 & 0.1072 & 0.7515 & 0.3969 \\
& vr\_room & 0.0956 & 0.0852 & 0.0904 & 0.8035 & 0.5453 \\
& 8b5caf3398 & 0.0618 & 0.0580 & 0.0599 & 0.8732 & 0.6045 \\
& b20a261fdf & 0.1103 & 0.1438 & 0.1270 & 0.8423 & 0.3553 \\
\cline{1-7}
\end{tabular}}
\label{tab:scanpp}
\end{table*}

\begin{table*}[t]
\centering
\footnotesize
\caption{Metrics of datasets used in our experiments.}
\resizebox{\linewidth}{!}{\begin{tabular}{c|c|c|c|c|c|c|c|c|c|c|c}
\cline{1-12}
Dataset & \multicolumn{4}{c|}{LERF-OVS~\cite{kerr2023lerf}} & \multicolumn{5}{c|}{MuSHRoom~\cite{ren2024mushroom}} & \multicolumn{2}{c}{ScanNet++~\cite{yeshwanth2023scannet++}} \\ \cline{1-12}
Scene   & figurines & ramen & teatime & waldo\_kitchen & coffee\_room & computer & honka & kokko & vr\_room & 8b5caf3398 & b20a261fdf \\ \cline{1-12}
Resolution  & 986 $\times$ 728  & 988 $\times$ 731 & 988 $\times$ 730 & 985 $\times$ 725 & \multicolumn{5}{c|}{738 $\times$ 994} & \multicolumn{2}{c}{1920 $\times$ 1440} \\ \cline{1-12}
Training Views  & 299 & 131 & 177 & 187 & 353 & 455 & 320 & 348 & 418 & 126  & 59  \\ \cline{1-12}
Initial points  & 65k & 27k & 23k & 15k & \multicolumn{5}{c|}{1000k} & 111k & 111k  \\ \cline{1-12}
\end{tabular}}
\label{tab:data}
\end{table*}

\end{document}